\documentclass[sigconf,authorversion,nonacm]{acmart}

\acmConference[ICSE 2022]{The 44th International Conference on Software Engineering}{May 21–29, 2022}{Pittsburgh, PA, USA}

\copyrightyear{2022} 
\acmYear{2022} 
\setcopyright{acmlicensed}\acmConference[ICSE '22]{44th International Conference on Software Engineering}{May 21--29, 2022}{Pittsburgh, PA, USA}
\acmBooktitle{44th International Conference on Software Engineering (ICSE '22), May 21--29, 2022, Pittsburgh, PA, USA}
\acmPrice{15.00}
\acmDOI{10.1145/3510003.3510112}
\acmISBN{978-1-4503-9221-1/22/05}

\AtBeginDocument{%
  \providecommand\BibTeX{{%
    \normalfont B\kern-0.5em{\scshape i\kern-0.25em b}\kern-0.8em\TeX}}}

\usepackage{comment}
\usepackage{graphicx}
\usepackage[T1]{fontenc}
\usepackage{listings}
\usepackage{xcolor}
\definecolor{verylightgray}{rgb}{.97,.97,.97}

\lstdefinestyle{Prologstyle}
{
    language=Prolog,
    basicstyle = \ttfamily\color{blue},
    moredelim = [s][\color{black}]{(}{)},
    literate =
        {:-}{{\textcolor{black}{:-}}}2
        {,}{{\textcolor{black}{,}}}1
        {.}{{\textcolor{black}{.}}}1
}

\usepackage{wrapfig}
\usepackage{booktabs}   
\usepackage{subcaption} 
\usepackage{amsmath}
\usepackage{hyperref}[]
\usepackage{tikz}
\usepackage[]{caption}

\usetikzlibrary{shapes, shapes.symbols, patterns, arrows, shapes, positioning,
	calc, fit, shadows, automata, shapes.geometric, decorations.text}
\usepackage[outline]{contour}
\contourlength{4pt}
\usepackage{multirow}
\def\inline{\lstinline[basicstyle=\normalsize\ttfamily,deletekeywords={Is,is}]}

\usepackage{pgfplots,pgfplotstable}
\usepgfplotslibrary{statistics}
\pgfplotsset{compat=1.8}

\usepackage{algorithm}
\usepackage[noend]{algpseudocode}
\algrenewcommand\Return{\State \algorithmicreturn{} }%

\algdef{SE}[DOWHILE]{Do}{doWhile}{\algorithmicdo}[1]{\algorithmicwhile\ #1}
\algnewcommand{\True}{\textbf{true}}
\algnewcommand{\False}{\textbf{false}}

\usepackage{enumitem}

\setitemize{noitemsep,topsep=0pt,parsep=0pt,partopsep=0pt, leftmargin=11pt}

\usepackage{soul}
\definecolor{darkgreen}{rgb}{0.0, 0.5, 0.0}
\newcommand{\rev}[1]{#1}

\usepackage{balance}

\begin{document}

\title{ExAIS: Executable AI Semantics}

\author{Richard Schumi}

\affiliation{%
  \institution{Singapore Management University}
  \country{Singapore}}
  \email{rschumi@smu.edu.sg}
  
\author{Jun Sun}
\affiliation{%
  \institution{Singapore Management University}
  \country{Singapore}}
  \email{junsun@smu.edu.sg}

\begin{abstract}
Neural networks can be regarded as a new programming paradigm, i.e., instead of building ever-more complex programs through (often informal) logical reasoning in the programmers' mind, complex `AI' systems are built by optimising generic neural network models with big data. In this new paradigm, AI frameworks such as TensorFlow and PyTorch play a key role, which is as essential as the compiler for traditional programs. It is known that the lack of a proper semantics for programming languages (such as C), i.e., a correctness specification for compilers, has contributed to many problematic program behaviours and security issues. While it is in general hard to have a correctness specification for compilers due to the high complexity of programming languages and their rapid evolution, we have a unique opportunity to do it right this time for neural networks (which have a limited set of functions, and most of them have stable semantics). In this work, we report our effort on providing a correctness specification of neural network frameworks such as TensorFlow. We specify the semantics of almost all TensorFlow layers in the logical programming language Prolog. We demonstrate the usefulness of the semantics through two applications. One is a fuzzing engine for TensorFlow, which features a strong oracle and a systematic way of generating valid neural networks. The other is a model validation approach which enables consistent bug reporting for TensorFlow models.
\end{abstract}

\keywords{AI frameworks, AI libraries, deep learning models, semantics, specification, test case generation, model validation, AI model generation}

\maketitle

\section{Introduction}
Artificial intelligence (AI) refers to intelligence expressed by machines. It can be based on imitating human intelligence or learning behaviour. For example, the technique called deep learning (DL) applies artificial neural networks to simulate the neurons of natural brains \cite{DBLP:journals/nn/Schmidhuber15}. In recent years, the dramatic growth of data and computing power has led to the spread of DL approaches which enabled numerous 
new technologies, like autonomous driving, language processing, and face recognition~\cite{DBLP:journals/csur/PouyanfarSYTTRS19}. 

In principle, AI systems can be seen as a new programming paradigm. Instead of building programs through (often informal) logical reasoning by programmers, AI systems are built by training neural network models with big data. 
\rev{For conventional programming, compilers or interpreters are major development tools. Similarly, for the development of neural networks the focus is on AI frameworks (or libraries) such as TensorFlow and PyTorch, which provide utilities, algorithms, and various types of layers that constitute DL models. 
}

\rev{The lack of a proper semantics (or correctness specifications) for programming languages has resulted in many critical issues such as inconsistencies or vulnerabilities.
Generally, it is hard to develop a comprehensive specification for compilers 
due to the high complexity and the rapid evolution of programming languages} \cite{DBLP:journals/iandc/SerbanutaRM09}.
\rev{However, we believe that it is more feasible for AI frameworks, since 
the semantics of AI frameworks are usually stable and simpler, i.e., the set of functions and layers used in popular machine learning models are usually limited.
}



Although AI systems have been proven successful in many applications, they are not error free. Especially in safety critical applications, like autonomous driving, or medical systems, even small bugs can have severe consequences. 
Thus, we need systematic well-grounded analysis methods for AI systems. While the lack of a proper language specification has hindered many program analysis tasks, we have here a unique opportunity to start afresh by building a formal semantics of AI systems, which then serves as a solid foundation for developing ever more sophisticated analysis methods for AI systems. In this work, we make such an attempt by developing an executable (formal) semantics for AI frameworks. That is, we introduce a Prolog specification for almost all TensorFlow layers. Our specification provides the foundation for solving many AI analysis problems, such as AI test generation, model validation, and potentially AI model synthesis. In the following, we present one example application of the semantics, AI test generation.

Recently, testing DL systems has become a popular research topic, with a variety of approaches~\cite{fehlmann2020autonomous,DBLP:journals/corr/abs-1906-10742,DBLP:journals/corr/abs-1806-07723}. In comparison, testing AI frameworks has gained less attention. 
A bug in an AI library can potentially put all systems that were built with the library at risk. Hence, it is crucial to ensure the correctness of these libraries. Multiple researchers made some early attempts~\cite{DBLP:conf/icse/DingKH17,DBLP:conf/issta/DwarakanathASRB18,DBLP:conf/icse/PhamLQT19,DBLP:conf/icml/SelsamLD17,DBLP:conf/icst/SharmaW19,DBLP:conf/aaai/SrisakaokulWAAX18,DBLP:conf/sigsoft/WangYCLZ20}, although existing approaches suffer from at least two major issues. First, existing approaches often have a weak oracle (i.e., a test case passes if some simple algebraic properties are satisfied~\cite{DBLP:conf/icse/DingKH17} or results in similar outputs on two implementations~\cite{DBLP:conf/icst/SharmaW19}), \rev{which can make it difficult to find deep semantic bugs.} 
Second, existing approaches fail to systematically generate valid AI models, i.e., to our knowledge, the state-of-the-art approach has a success rate of 25\% in generating valid models (after a complicated process combining NLP techniques with manually designed templates)~\cite{li2020documentation}. 


Our executable semantics solves these two issues naturally. 
First, given an AI model, our semantics can be executed to compute precisely the constraints that are satisfied by the output (based on the Prolog's underlying unification and rewriting system), which serves as a strong test oracle. 
Second, our semantics facilitates the development of a fuzzing-based test generation method to systematically produce valid AI models. Our semantics facilitates the generation process by providing feedback in the form of a badness-value that indicates how far the model is away from being valid (i.e., satisfying the precondition of all layers of the model). Our evaluation shows that we significantly improve the success rate of generating valid AI model (e.g., 99\%) and \rev{can generate diverse model architectures consisting of various layers (in the form of sequences and complex graphs) as well as layer arguments, weights, and input data.}

Besides, having an executable semantics of an AI library opens the door for a range of other exciting research. For instance, it allows systematic  validation and bug reporting for AI models, e.g., by investigating automatically whether certain preconditions required by a specific AI component are violated, \rev{and by checking if the architecture of a model is valid.} We remark that existing debugging techniques for AI systems rely on simple oracles (such as shape mismatch), which is preliminary compared to the oracle that is provided by our semantics. For another instance, our semantics potentially supports the development of synthesising techniques for AI models. That is, given certain user requirements (e.g., on the input format, the output constraints and perhaps a model sketch), the problem of synthesising AI models can be reduced to a search problem based on the well-specified semantics of all components in existing AI frameworks, i.e., by composing the components in different ways and checking if the user requirements are satisfied.

To sum up, we present the following contributions in this work. 
\begin{itemize}
    \item We introduce an executable semantics of an AI library (i.e., TensorFlow), which captures the behaviour of the DL layers.
    \item To demonstrate the relevance of our semantics, we present a novel testing method for AI libraries that applies the semantics as test oracle and utilises it to generate valid AI models. \rev{Our method was able to reveal 14 issues and bugs in TensorFlow.}
    \item We illustrate a model validation method that can identify issues of invalid AI models, \rev{like a wrong model architecture}, based on our semantics, and that supports consistent bug reporting.
\end{itemize}

\paragraph{Structure.} The rest of the paper is structured as follows. In Sect.~\ref{sec:spec}, we introduce our semantics and show how we specify different layers.
In Sect.~\ref{sec:method}, we present two applications of the semantics. In Sect.~\ref{sec:evaluation}, we evaluate these two applications systematically. 
Lastly, we review the related work in Sect.~\ref{sec:related} and conclude in Sect.~\ref{sec:conclusion}.

\section{AI Framework Semantics}\label{sec:spec} 
In this section, we start with a brief introduction of the logic programming language Prolog~\cite{nilsson1990logic} that we adopt for the development of our AI framework semantics. Afterwards, we describe our approach on developing the semantics through examples. 
In general, to support a variety of analysis tasks including testing, debugging and even synthesising AI models, it is important that the semantics fulfils the following requirements \cite{DBLP:journals/jlp/RosuS10,DBLP:journals/iandc/SerbanutaRM09}. First, the semantics must be declarative and high-level so that it avoids complications due to implementation-level optimisation and to facilitate a concise specification that is independent of implementation details. \rev{Second, it must support symbolic reasoning to allow various property, consistency and correctness checks, and thus it must be specified in logic. Lastly, it must be executable so that it can be used to support a variety of automated analysis tasks, like testing or debugging.}


\subsection{Prolog}
Prolog is a declarative language that relies on first order logic. Programs in Prolog are built with rules and facts, \rev{which are usually concerned with mathematical relations. Hence, we believe that Prolog is suitable for a formal semantic representation. There might be more formal mathematical semantic models that, e.g., rely on} Coq~\cite{DBLP:conf/icml/SelsamLD17}, \rev{but such notations are not as flexible and would make it difficult to support a large number of layers.}

In Prolog, a user can ask queries, and the answer is computed automatically based on the rules and facts with a unification algorithm.
The following example shows two facts stating that \inline|average| and \inline|flatten| are layers and a rule that says that layers are AI components \inline|ai_components|.
\begin{lstlisting}[language=Prolog, numbers=none]
layer(average).
layer(flatten).
ai_components(X):- layer(X).
\end{lstlisting}
A user may ask queries such as \inline|ai_components(flatten).|, which is simply answered by true, or \inline|ai_components(X).| where $X$ is a variable which produces the following results.
\begin{lstlisting}[language=Prolog, numbers=none]
X = average; 
X = flatten.
\end{lstlisting}
A rule can generally be seen as a predicate that has arguments (in the header) and describes the relation between these arguments with a conjunction of clauses in the body of the rule. Arguments can be  atoms (lowercase) or variables (uppercase). Prolog supports a number of  data types and has built-in features for various mathematical operations. Moreover, it includes functions for list and string processing, and higher order programming notations~\cite{DBLP:books/sp/Bramer13}.

Prolog is a well-suited specification language for our purpose for multiple reasons. First, it is largely intuitive and convenient to use (i.e., by specifying rules and facts) and has a reasonably large user base, which is extremely important as we intend to invite the open-source community to collaboratively maintain the semantics. Second, it supports various list operations and mathematical expressions, which is helpful for capturing the semantics of AI libraries that involve mathematical operations on multidimensional data. Third, Prolog supports automated reasoning (through an underlying rewriting and search system), which is essential for our purpose. Lastly, Prolog is a well-studied declarative programming language, which is fundamentally different from the imperative programming that is used to implement AI frameworks. This makes it unlikely that our semantics will have the same bugs in the implementation. 

\subsection{Specifying the Semantics}
In this work, we focus on developing a semantics for the TensorFlow framework, since it is the most popular AI framework. TensorFlow is a complex framework that provides various tools, implementations, and features for the development of AI systems. We systematically go through every component in the TensorFlow framework and identify all the components which are relevant to AI models themselves (\rev{rather than, components for the training process}). Our goal is to implement a comprehensive semantics for one of its core components, i.e., the layers~\cite{layerdocs}. We specify the semantics of nearly all layers except a few layers for practical reasons. A detailed discussion of the omitted layers is in Sect.~\ref{sec:evaluation}.
\rev{In total, we specify the layer functionality that is required to perform a prediction of 72 layers of different types, we did not implement any TensorFlow algorithms that are concerned with the training of models or data processing.}
A full list of these layers is available in our repository~\cite{repo}.
For each layer, we systematically go through the documentation, and run tests to understand the semantics. Unfortunately, the semantics of quite a number of layers are under-specified,
and we have to seek further help from online forums and/or the authors. In multiple cases, we identify under-specified or mis-specified behaviours, like wrong handling of unexpected layer arguments, or missing restrictions for layer inputs (more details are in Sect.~\ref{sec:evaluation}). This suggests that formalising the semantics alone is already useful, and can help to improve AI frameworks. 

\begin{figure}[t]
	\centering
	\begin{tikzpicture}
	\begin{axis}[width=\linewidth,height=12em,ybar,ymin=0,    ylabel={frequency},
	xlabel={line count}]
	\addplot +[hist={bins=20}] table [y index=0] {linecount.csv};
	\end{axis}
	\end{tikzpicture}
	\caption{Histogram of the frequencies of line numbers used to specify the layers in Prolog.}
	\label{fig:linecount}
\end{figure}
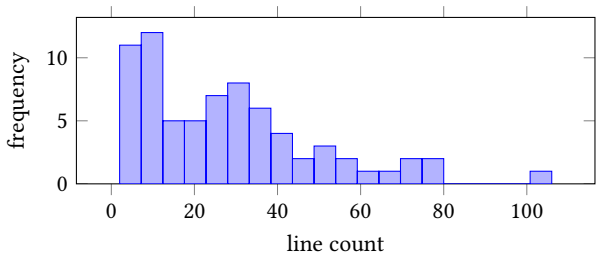

The semantics of each layer is specified using one or more rules in Prolog. Figure~\ref{fig:linecount} shows a histogram of the frequencies of the number of lines (or rules) required for the layers. It can be seen that many layers only require less than 10 lines. The reason is that they reuse predicates of other layers, or built-in rules. A major fraction of layers requires no more than 40 lines, and the remaining layers have nearly all less than 80 lines. One outlier (the Dot layer) has 106 lines since the implementation supports various axis combinations for the dot product. In total, the semantics consists of about 3200 lines of Prolog code and on average 44 lines per layer (when shared predicates are considered).
The implementation of nearly all layers has an one-to-one correspondence with the deterministic functionality of the TensorFlow layers. This allows us to, given an AI model, straightforwardly invoke the relevant rules in our semantics and apply them to generate constraints on the model output.

We remark that since these Prolog rules are specified by humans, they are potentially buggy as well. Standard practices are applied to ensure the correctness of the semantics itself, such as code review, manual testing (e.g., using documentation examples) and automated testing (e.g., using our own fuzzing engine). While we cannot be completely sure that the semantics is correct, we have reason to believe that it is reasonably so. In general, one can always argue that correctness is an illusion and all there is is consistency (e.g., between manually-programmed TensorFlow and manually-written Prolog semantics). That is, as long as the same bug is not present in both 
the implementation and the semantics, 
inconsistency would arise and either TensorFlow or our semantics will be fixed. 
In the following, we illustrate our semantics with multiple examples.

\paragraph{Dense layer.} Listing~\ref{lst:dense} shows our Prolog semantics of a Dense layer~\cite{dense}. It is a standard densely connected layer, which contains a number of nodes and each is connected with all inputs. 
The output is computed by the weighted sum of all inputs at each node with an added bias value.
%
Our specification works as follows. The rule starting in Line~1 has multiple parameters: a list \inline{[I|Is]}, a weight array \inline|IWs|, and a bias array \inline|Bs| (which can be intuitively regarded as inputs) and a list \inline|Os| (which can be regarded as the output). The list notation \inline{[I|Is]} enables access to the first element \inline{I} and the remaining elements \inline{Is} of a list. Line~2 constrains the depth of the  nested input to two. We handle higher dimensional input separately.
Line~3 applies a predicate that is true when \inline{O} can be unified as the expected output for an input \inline{I},
and Line~4 is a recursive constraint, which intuitively continues with the next inputs.

The rule in Lines~6--9 is similar, except that it handles (layer) inputs with a higher dimension, which is checked in Line~7, and recursively uses our initial predicate from Line~1 since the dense layer only performs computations in the inner most list even when it receives high dimensional input data.
Line~11 (and Line~18) are the base cases for the recursion, i.e., when only an empty list remains.

The predicate in Line~13  encodes the main layer functionality and becomes true when the \inline|Res| variable is the expected output for the input \inline{[I|Is]}.
It has the same arguments as our first rule and an additional temporary variable \inline|Res0| for the result. 
It consists of clauses for multiplying the weight arrays \inline|IW| with each input \inline|I| and for adding the results in Line~16. The predicates \inline|multiply_list_with| and \inline|add_lists| are straightforward and are therefore omitted.

\begin{lstlisting}[language=Prolog,  xleftmargin=15pt, float=tp,
%aboveskip=-6pt,belowskip=-6pt, 
caption={Prolog semantics of the Dense layer.},
 label={lst:dense},deletekeywords={is}, morekeywords={dense_layer,dense_node_comp,depth,add_lists,multiply_list_with,padding1D}]
dense_layer([I|Is], IWs, Bs, [O|Os]) :-
    depth([I|Is],2),
    dense_node_comp(I, IWs, Bs, O),
    dense_layer(Is, IWs, Bs, Os).
    
dense_layer([I|Is], IWs, Bs, [O|Os]) :-
    depth([I|Is],D), D > 2,
    dense_layer(I, IWs, Bs, O),
    dense_layer(Is, IWs, Bs, Os).
    
dense_layer([], _, _, []).

dense_node_comp([I|Is],[IW|IWs],Res0,Res) :-
    multiply_list_with(IW,I,Res1),
    add_lists(Res0,Res1,Res2),
    dense_node_comp(Is,IWs,Res2,Res).
    
dense_node_comp([],[],Res,Res).
\end{lstlisting}
With this Prolog semantics, we can now answer a variety of queries, e.g., to compute the expected output of a Dense layer. The following listing shows some examples. The first is a basic query with two inputs and two nodes and the second has three nodes and more complex input.
It can be seen that even such a small and simple specification can handle high dimensional input with just a few lines of code. 
The third query fails due to invalid arguments since the number of nodes must be equivalent to the size of the bias array and also to the second dimension of the weight array. Many layers have non-trivial preconditions like that. This makes the generation of test data difficult. In Section~\ref{sec:method}, 
we will illustrate how we can identify such invalid inputs and produce useful information about the source and severity of such issues to support our model validation and test generation approaches. 
\begin{lstlisting}[language=Prolog,  numbers=none, %xleftmargin=20pt, 
 morekeywords={dense_layer,sig_gate,depth,add_lists,multiply_list,tanh_gate}]
dense_layer([[4,9]], [[5,8],[7,6]], [4,3], X).
X = [[87, 89]]

dense_layer([[[[[1,2,3]]],[[[4,5,6]]]]], [[2,3,4],[5,4,6],[7,6,8]], [2,3,4], X).
X = [[[[[35, 32, 44]]], [[[77, 71, 98]]]]] 

dense_layer([[3,1]], [[8,4],[2,7],[3,9]], [7], X).
false.
\end{lstlisting}

\paragraph{Conv1D layer.} 
A convolution layer~\cite{conv} shifts a kernel (or pool) over the input space and produces the weighted sum of the elements within the kernel. It can, for instance, be used for image preprocessing or recognition.
Listing~\ref{lst:spec} illustrates the (simplified for the sake of presentation) semantics of this layer. 
Line~1 shows a predicate for the layer, which has the (layer) input data \inline|Is|, a number \inline|KernelSize| for the one dimensional size of the window that should be shifted, a weight array \inline|IWs| for the kernel elements, a bias array \inline|Bs| of values that are added to the output,  a step size \inline|Strides|, a Boolean \inline|Padding| for optionally enlarging the input data, e.g., by surrounding it with zeros, and the output data \inline|Os|, as parameters. 
\rev{Lines~2--4 apply predicates for checking the validity of the layer inputs and arguments, i.e., to ensure that the input dimensions are as required, to check that the kernel is not too large for the input, and to verify that the weights are valid. An example implementation of such a precondition is given in} Sect.~\ref{sec:modelvalidation}.
Line~5 applies the \inline|pool1D| predicate, which is a generic predicate for different types of pooling or convolutional operations, and it becomes true when \inline|Os| can be unified with the result of the operation.


\begin{lstlisting}[language=Prolog,  xleftmargin=15pt, float=tp, 
%aboveskip=-6pt,belowskip=-5pt, 
caption={Simplified Prolog specification of a pooling layer.}, label={lst:spec},deletekeywords={is}, morekeywords={pool1D,conv1D_layer,calc_padding,insert_pool_field,get_pool_res1D,padding1D,false}]
conv1D_layer(Is,KernelSize,IWs,Bs,Strides,Padding,Os):- 
	check_dimensions(Is,3),
	check_valid_kernel(Is,KernelSize,Padding),
	check_valid_weight_shapes(Is,KernelSize,IWs,Bs),
	pool1D(sum,Is,KernelSize,Strides,Padding,IWs,Bs,false,Os).

pool1D(Poolfunc,[I|Is],PoolSize,Strides,Padding,IWs,Bs,MultiLayerPool,[O|Os]):- 
	pool1D(Poolfunc,I,0,0,PoolSize,Strides,Padding,IWs,Bs,MultiLayerPool,[],O),
	pool1D(Poolfunc,Is,PoolSize,Strides,Padding,IWs,Bs,MultiLayerPool,Os).
	
pool1D(_,[],_,_,_,_,_,_,[]).

pool1D(Poolfunc,[[I|Is0]|Is],0,0,PoolSize,Strides,true,IWs,Bs,MultiLayerPool,[],Os) :-
	atomic(I),length([[I|Is0]|Is],L),
	calc_padding(L,PoolSize,Strides,LeftP,RightP),
	padding1D([[I|Is0]|Is], x,LeftP, RightP, Is1),
	pool1D(Poolfunc,Is1,0,0,PoolSize,Strides,false,IWs,Bs,MultiLayerPool,[],Os).
	
pool1D(Poolfunc,[[I|Is0]|Is],X,0,PoolSize,Strides,false,IWs,Bs,false,Os0,Os) :-
	atomic(I),length([[I|Is0]|Is],LX), 
	get_pool_res1D(Poolfunc,[[I|Is0]|Is],X,Y,PoolSize,Strides,IWs,Bs,false,O),
	insert_pool_field(Os0,O,true,X,Y,Strides,Os1),
	(X+Strides+PoolSize =< LX -> X1 is X+Strides; X1 is LX+1),
	pool1D(Poolfunc,[[I|Is0]|Is],X1,0,PoolSize,Strides,false,IWs,Bs,false,Os1,Os).

pool1D(Poolfunc,[[I|Is0]|Is],X,Y,PoolSize,Strides,Padding,IWs,Bs,true,Os0,Os) :-
	atomic(I),length([[I|Is0]|Is],LX), 
	get_pool_res1D(Poolfunc,[[I|Is0]|Is],X,Y,PoolSize,Strides,IWs,Bs,true,O),
	insert_pool_field(Os0,O,true,X,Y,Strides,Os1),
	(X+Strides+PoolSize =< LX -> X1 is X+Strides,Y1 is Y; X1 is 0,Y1 is Y+1),
	pool1D(Poolfunc,[[I|Is0]|Is],X1,Y1,PoolSize,Strides,Padding,IWs,Bs,true,Os1,Os).

pool1D(_,[[I|Is0]|Is],X,Y,_,_,false,_,_,_,Os,Os) :-
	atomic(I),
	(length([[I|Is0]|Is],LX), X >= LX; 
	length([I|Is0],LY), Y >= LY).
\end{lstlisting}
Line~7 shows the \inline|pool1D| predicate, which has the same arguments as the 
\inline|conv1D_layer| with an additional \inline|Poolfunc| meta-predicate to decide what should happen 
within the pool, and an argument \inline|MultiLayerPool| to specify if the pool should be shifted over the second axis of the input space.
Next, there is a call to a sub-predicate for one sample of the input in Line~8, with initialisations of coordinates and a temporary output variable, 
Line~9 iterates over the other samples, and Line~11 is a simple stopping clause.

In Lines~13--17, we apply padding with the \inline|calc_padding| predicate that is true when \inline|LeftP|, and \inline|RightP| are padding sizes to maintain the input shape after pooling, and by using these variables for the \inline|padding1D| predicate that succeeds when \inline|Is1| is \inline|Is| with padding.
The predicate in Lines~19--24 does the actual application of the convolution.
Line~21 utilises the pool predicate, i.e. the weighted sum of the kernel, for a given position \inline|X,Y|, which is then appended to the output in Line~22. 
Next, Line~23 has a conditional clause that introduces new variables \inline|X1,Y1| for the next coordinates given the current pool position, and finally Line~24 performs recursion to continue with the next coordinates. 

Lines~26--31 are defined similarly, except that the pool is also shifted over the second axis by iterating over the \inline|Y| coordinate instead of just using the sum over these values.
Finally, Lines~33--36 show a stopping clause, which becomes true when one of the indexes \inline|X,Y| is outside the input space.

Similarly, with the above semantics, we can now query this layer as shown in the following examples. The first example has integer and the second has floating-point inputs. It can be seen that variables are not limited to a specific type, our implementations works for integers as well as floating-point numbers.
\begin{lstlisting}[language=Prolog,  numbers=none, %xleftmargin=20pt,
 morekeywords={dense_layer,sig_gate,depth,add_lists,multiply_list,tanh_gate,conv1D_layer}]
conv1D_layer([[[9, 9, 6, 5, 3], [4, 5, 5, 8, 2], [2, 4, 2, 3, 10]]], 3,[[[4, 5], [4, 5], [3, 4], [2, 5], [4, 3]], [[5, 4], [5, 1], [3, 1], [5, 2], [3, 5]], [[2, 4], [1, 1], [5, 3], [3, 1], [3, 5]]],[0, 0], 1, false, X).
X = [[[275, 271]]] 

conv1D_layer([[[0.0113, 0.1557, 0.1804], [0.8732, 0.317, 0.9175], [0.7246, 0.833, 0.8881]]], 2,[[[0.0419, 0.2172], [0.9973, 0.6763], [0.6917, 0.452]], [[0.0743, 0.9004], [0.52, 0.5426], [0.4529, 0.5032]]],[0, 0], 1, false, X).
X = [[[0.92579027, 1.60921455], [1.8765842, 2.3700637]]]
\end{lstlisting}
An interesting aspect of this layer is that a lot of the functionality can be reused for other layers. In particular, the \inline|pool1D| predicate is defined generic so that it can be applied for the semantics of many related layers. For example, it is applied for simple pooling layers, like (global) average or max pooling and for other convolutional layers, like separable convolution and locally connected layers.

\paragraph{Non-deterministic layers.} 
Although most layers have deterministic behaviour when their weights are known, there are a few exceptions including layers such as Dropout, AlphaDropout, SpatialDropout, GaussianDropout, and GaussianNoise. All of them either have built-in non-determinism (i.e., random behaviour) or are probabilistic in nature (i.e., producing outputs that follows a certain probability distribution).
We defined Prolog rules that constrain these layers to satisfy high-level properties which are documented in the respective documentation, e.g., if the layer is expected to produce results that are according to a specific distribution, they should. We demonstrate how this is done for the dropout layer~\cite{dropout}. This layer is expected to set values of the input to zero according to a given rate. \rev{(Additionally, the layer also adds noise to the data, but we neglect this feature in the example for brevity.)}
Listing~\ref{lst:dropout} shows a predicate that checks if the dropout actually occurs at a given rate. The arguments are the inputs \inline|Is|, the outputs from TensorFlow \inline|Os|, the rate, and a threshold \inline|AcceptedRateDiff| for the maximum difference to the observed rate.
\rev{Line~2 shows predicates which are true when} \inline|N| and \inline|NO| \rev{can be unified with the number of input and output values, which must be equal.}
Line~3 does the same for the number of zeros in the input, and Line~4 for the number of zeros in the given output.
Next, Line~5 sets \inline|RealRate| to the observed rate of dropouts while considering the already existing zeros in the input, 
and Line~6 applies the \inline|abs| function to obtain the difference to the expected rate.
\rev{Finally, Line~7 checks if the difference is bigger than our threshold and writes a message if that is the case.} 

\begin{lstlisting}[language=Prolog,  xleftmargin=15pt, float=tp, 
%aboveskip=-6pt,belowskip=-5pt, 
caption={Prolog specification of the Dropout layer.}, label={lst:dropout},deletekeywords={is}, morekeywords={lstm_layer,dense_node_comp,depth,add_lists,multiply_list_with,padding1D,dropout_layer}]
dropout_layer(Is, Os, Rate, AcceptedRateDiff) :- 
	count_atoms(Is,N), count_atoms(Os,NO), NO = N,
	count_occurrences(Is,0,NZeroOrig), 
	count_occurrences(Os,0,NZeroNew),
	RealRate is (NZeroNew - NZeroOrig) / (N -NZeroOrig),
	Diff is abs(Rate - RealRate),
	(Diff > AcceptedRateDiff -> (write("Expected Rate: "), writeln(Rate), write(" Actual Rate: "), writeln(RealRate),  false); true).
\end{lstlisting}


\section{Applications of the Semantics}\label{sec:method} 
The semantics has many applications. In this section, we demonstrate two of them. We illustrate the overall design of our testing and model validation  approaches and show step-by-step examples. 

\subsection{AI Framework Testing}
Recently, multiple researchers started working on AI framework testing~\cite{DBLP:conf/icse/DingKH17,DBLP:conf/issta/DwarakanathASRB18,DBLP:conf/icse/PhamLQT19,DBLP:conf/icml/SelsamLD17,DBLP:conf/icst/SharmaW19,DBLP:conf/aaai/SrisakaokulWAAX18,DBLP:conf/sigsoft/WangYCLZ20}. While an impressive number of bugs have been identified, there are still many open challenges. Based on our executable semantics, we introduce a novel AI framework testing method which improves the existing approaches by addressing the following problems.
(1) The oracle problem (i.e., how we verify if the output is correct). 
(2) The test generation problem (i.e., how can we effectively generate valid test cases).
For the first problem, the existing approaches focus on differential testing~\cite{DBLP:conf/icse/PhamLQT19,DBLP:conf/aaai/SrisakaokulWAAX18,DBLP:conf/sigsoft/WangYCLZ20} or a simple form of metamorphic testing~\cite{DBLP:conf/icse/DingKH17,DBLP:conf/issta/DwarakanathASRB18,DBLP:conf/icst/SharmaW19}. 
For differential testing, multiple libraries (or implementations) are compared against each other, and bugs are identified when an inconsistency is found. 
The kind of metamorphic testing done on AI frameworks introduces changes in the test input (i.e., an AI model) that should not affect the output. A different output would thus indicate a bug. In other words, such testing relies on the algebraic property that `dead code' does not change the model's behaviour. 
Differential testing is powerful since a second implementation can serve as a good test oracle. However, a second independent implementation is not always available, especially for new algorithms. Even if it is available, it is still possible that it suffers from the same bugs due to reasons such as using the same underlying library or implementing certain optimisation in the same `wrong' way. 
\rev{Our method can also be seen as a form of differential testing, but 
it is unlikely that our semantics will suffers from the same bugs since it is developed in a different modelling paradigm at a high-level (e.g., where little performance optimisation is involved).} On the other hand, metamorphic testing is usually limited to specific types of bugs based on algebraic properties (i.e., a very small piece of the semantics) which are used to guide the test generation. Another issue that limits the applicability of both approaches is the non-deterministic behaviour of the layers, e.g., inconsistency may be due to non-deterministic or probabilistic behaviours rather than bugs. Usually, the training of AI models contains random factors that cause small discrepancies in different implementations and make an exact comparison and the identification of bugs difficult. This issue is sometimes resolved by relying on approximating oracles that accept a range of inputs. This is not an ideal solution since it often requires manual adjustments~\cite{DBLP:conf/kbse/NejadgholiY19}. 
Based on our semantics, we have an alternative more powerful solution for this problem. By developing our semantics with a focus on the high-level behaviour of the layers, we can test the results of an AI model independently of the non-deterministic training algorithms. This facilities the detection of subtle bugs without the interference of underlying non-deterministic or probabilistic behaviours. 

The second problem, i.e., the test case generation problem, is often addressed by using existing benchmark examples or by modifying these examples, e.g., via mutation~\cite{DBLP:conf/sigsoft/WangYCLZ20}. The problem of the former is that existing examples are limited in both numbers and variety. The problem of the latter is that test cases generated through mutation are often invalid, e.g., due to non-trivial preconditions that must be satisfied by the layers. Most AI libraries contain dozens of different layers that can be connected in sequences or in complex graph structures to form  deep learning models. There are simple layers, like a densely connected layer, and also highly complicated layers which apply operations on multidimensional inputs or even combine the functionality of other layers. Usually, they need a number of configuration arguments, e.g., for specifying the number of neurons or a filter size. Most layers have preconditions for the inputs and hence they cannot be freely combined. To address this problem, a recent approach~\cite{li2020documentation} proposes to extract information about the input requirements from the library documentation and to generate `valid' test cases accordingly. However, after all its effort, it still has very limited capability in producing valid test data, i.e., the success rate is around 25\%. 

\begin{figure}[t]
\tikzstyle{io} = [trapezium,trapezium left angle=70,trapezium right
angle=-70, draw, minimum height=3.5em, text width=5em, text centered]
\tikzstyle{block} = [rectangle, draw, fill=gray!30, thick,
text width=5.7em, text centered, rounded corners, minimum height=4.6em]
\tikzstyle{plus} = [circle,  draw, fill=gray!30, thick, text centered]
\tikzstyle{line} = [draw, -latex, thick]
   	\centering
   	\resizebox{0.99\linewidth}{!}{
   		\begin{tikzpicture}[node distance=11em, auto, font=\sffamily\large]
   		
   		
   		\node [block] (tcg) {Test Case Generation};
   		\node [io, text width=6em, right of= tcg] (spec) {Prolog Specification};
   		\node [io, text width=4.2em, left of= tcg] (lib) {AI\\ Library};
   		\node [io, text width=4.2em, below=3em of tcg] (tc) {Test Cases};
   		\node [block, left of=tc] (aiex) {Execution with AI Lib};
   		\node [block, right of=tc] (specex) {Execution with Specification};
   		\node [io, below=3em of aiex] (o1) {Test Outputs};
   		\node [io, below=3em of specex] (o2) {Test Outputs};
   		\node [block, below=3em of tc] (com) {Output Comparison};
   		\node [io, text width=4.4em, inner xsep=0, below=3em of com] (ver) {Pass/Fail Verdict};
   		
   		\path [line] (lib)++(0,1) -- node {}(lib);
   		\path [line] (lib)-- node {}(aiex);
   		\path [line] (spec)++(0,1) -- node {}(spec);
   		\path [line] (spec)-- node {}(specex);
   		\path [line] (spec)-- node {}(tcg);
   		\path [line] (tcg)-- node {}(tc);
   		\path [line] (tc)-- node {}(aiex);
   		\path [line] (tc)-- node {}(specex);
   		\path [line] (aiex)-- node {}(o1);
   		\path [line] (specex)-- node {}(o2);
   		\path [line] (o1)-- node {}(com);
   		\path [line] (o2)-- node {}(com);
   		\path [line] (com)-- node {}(ver);

		\matrix [draw, right of=ver]  {
		  \node [io,minimum height=1.6em, text width=-0.1em,label=right:{}] (l1){};
		  \node [right=0.2em of l1,minimum height=1.8em, text width=4.6em] {\small input/output \\[-0.3em] artefact}; \\
		  \node [block,minimum height=1.8em, text width=1.4em,label=right:{\small process}] {}; \\
		};
   		\end{tikzpicture}
   	}
   	\caption{\rev{Overview of the data flow of our testing approach.}}
   	\label{fig:overview}
\end{figure}
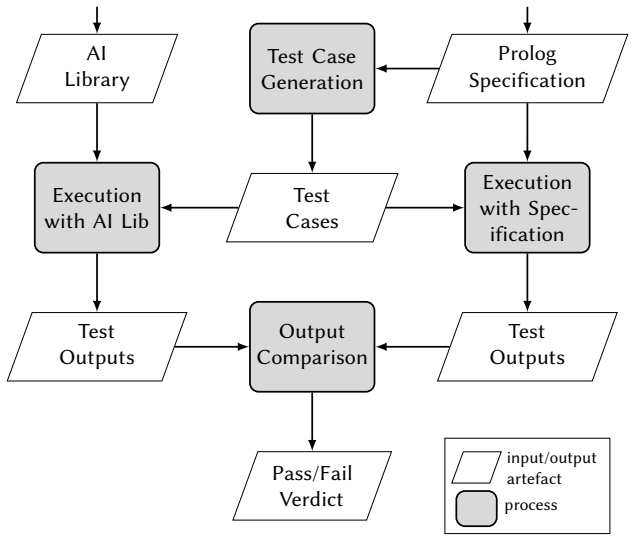
In this work, we develop a fuzzing method which utilises feedback from our semantics to enable a systematic way of producing valid test cases, i.e., by finding layers (and associated arguments) that are compatible with each other. Moreover, our approach explores various layer arguments in order to find interesting combinations that might reveal bugs.
The overall workflow is depicted in Fig.~\ref{fig:overview}. Next, we will explain the involved tasks and components.

Our method has two main components. First, the most important component is the oracle, which is our executable Prolog semantics. Given a deep learning model (in the form of collection of layers and their arguments) and the input data, the oracle executes the semantics to produce a prediction, in a similar fashion as it is done by an AI library. 
Second, the test case generator is a tool that produces test data in the form of multidimensional arrays and a deep learning model as a collection of layers and their respective arguments. Our test case generator is based on fuzzing techniques and produces random input data and models. The focus of the generation is to produce complex models with graph structures in order to extensively test combinations of different layers in various settings. 


In the following, we use the simplified semantics of the convolution layer that we showed in the last section to demonstrate the test generation. \rev{The test case generator produces test input data as well as a deep learning model (including layer arguments) to reveal bugs or inconsistencies in the AI library. The model and the input data (for performing a prediction) together form a test case.} 
A test case is automatically generated in the form of a Prolog query, which is given to the Prolog interpreter to compute a result with a unification algorithm. 
For example, a simple query for our convolutional layer is as follows.
 \begin{lstlisting}[language=Prolog,numbers=none,morekeywords={conv1D_layer,true, false}]
 conv1D_layer([[[5,8,7],[9,10,7],[3,7,7]]],2, [[[1],[1],[1]],[[1],[1],[1]]],[0],1,false,1,X).
 \end{lstlisting}
The same test case for TensorFlow is expressed as a simple Python program as shown in Listing~\ref{lst:tensorflow}. The first lines are imports and configurations. Lines~3-4 define the actual deep learning model. In Lines~5-8 we overwrite the weights for the kernel and the biases in order to support a deterministic prediction. Usually these weights are initialised randomly and tuned through training. Line~9 sets the input data and Line~10 performs a prediction with the model for the input data, which gives us the test output.

\begin{lstlisting}[language=Python,  xleftmargin=15pt, float=tp, 
%aboveskip=-5pt,belowskip=-7pt, 
caption={TensorFlow example of a Conv1D layer.}, label={lst:tensorflow}]
import tensorflow as tf, numpy as np
from tensorflow.keras import layers, models
model = tf.keras.Sequential([
    layers.Conv1D(1, (2),strides=(1), input_shape=(3, 3)) ])
w = model.get_weights()
w[0] = np.array([[[1], [1], [1]], [[1], [1], [1]]])
w[1] = np.array([0])
model.set_weights(w)
x = tf.constant([[[5, 8, 7], [9, 10, 7], [3, 7, 7]]])
print (np.array2string(model.predict(x,steps=1)))
\end{lstlisting}
Finally, we compare the output of the test program (that includes the AI library) to the output of the semantics. A test fails if the constraints produced by the semantics are not satisfied by the output from the Python program. 
In general, such a failure can be caused by bugs in the library and possibly by bugs in the semantics.
For our example, the output was \inline|[[[46], [43]]]| in both cases, which represents a successful test.
\rev{A bug is identified when there is an inconsistency in the outputs (prediction results) of the TensorFlow program and of our semantics. If there is an inconsistency, we localise the bug by checking which layer specification is violated, and by manually checking the cause.}


\subsection{Test Case Generation}
The test case generator is a tool that can produce models in the form of Python scripts \rev{(that use TensorFlow)} as well as models represented as Prolog queries.
\rev{It also generates random test inputs (in the form of multidimensional arrays that need to have a specific shape or number of dimensions depending on the first layer) for a prediction with a model and layer arguments, which can also be the weights of layers.
The generation is done by the test case generator alongside the AI model generation while considering input requirements. 
The details about the required arguments can be found in the TensorFlow documentation. A test case is a model together with the associated test inputs.}

\begin{algorithm}[t]
	\caption{Pseudo code of the test model generation.}
	\label{alg:testgeneration}
	\scriptsize
	\begin{algorithmic}[1]
		\Require $\mathit{GenHelper}$: helper class containing a map of layer generators  (Layer.Name → Generator)
				$\mathit{SemanticHelper}$ helper class for the execution of the semantics 
		\Function{recursiveGeneration}{level, rand}
		\If{$\mathit{level = 0 \lor rand.nextBool()}$}
		\Comment{Allow deeper graph based on the random result}
			\Return $\mathit{null}$ \Comment{End of the recursion}
		\EndIf
		\State 	$\mathit{layer \gets GenHelper.generateLayer()}$ \Comment{Generates a random layer incl. arguments}
		\State 	$\mathit{inputNumber} \gets 1$
		\If{$\mathit{layer.hasMultipleInputs()}$} \Comment{Check if layer requires multiple inputs}
			\State $\mathit{inputNumber \gets rand.nextInteger(3)}$ 
		\EndIf
		\For{$\mathit{i \gets 1}$ {\bf to} $\mathit{inputNumber}$}
			\State $\mathit{layer1 \gets recursiveGeneration(level-1,rand)}$ 
			\State $\mathit{layer1.setParent(layer), layer.addChild(layer1)}$
			\Comment{Connect layer}
		\EndFor%
		\Return $\mathit{layer}$
		\EndFunction
		
		\Function{findValidModel}{layer, maxTries, rand}
		\Comment{find valid model given the random model}
		\For{$i \gets 1$ {\bf to} $maxTries$}
			\State $\mathit{(success, error) \gets SemanticHelper.run(layer)}$\Comment{run the semantics, get success/error}
			\If{$success$}
				\Return $\mathit{layer}$
			\ElsIf{$\mathit{error.getLocation() = lastError.getLocation() \, \land}$ \newline \hspace*{5.8em} $\mathit{error.getBadness() \geq lastError.getBadness()}$}
				\State $\mathit{layer \gets lastLayer, error \gets lastError}$ 
				\Comment{Restore model if no improvement}
			\Else
				\State $\mathit{lastLayer \gets layer, lastError \gets error, layer1 \gets error.getLocation() }$ 
				\If{$rand.nextBool()$}
					\State $\mathit{layer1.regenerateArgs()}$ \Comment{Reset the layer arguments}
				\ElsIf{$rand.nextBool()$}
					\State $\mathit{options \gets ["Reshape","UpSampling","Cropping","Padding",\ldots]}$
					\State $\mathit{layer2 \gets GenHelper.generateLayer(options)}$ \Comment{Generate one of the options}
					\State $\mathit{layer1.insertLayerBeforeChild(layer2)}$\Comment{Insert layer before input}
				\ElsIf{$rand.nextBool()$}
					\State $\mathit{layer2 \gets GenHelper.generateLayer()}$
					\State $\mathit{layer1.replaceWith(layer2)}$\Comment{Replace current layer with another layer}
				\ElsIf{$rand.nextBool()$}
					\State $\mathit{layer2 \gets GenHelper.generateLayer()}$
					\State $\mathit{layer1.replaceChildWith(layer2)}$\Comment{Replace input layer with another layer}
				\EndIf
			\EndIf
		\EndFor
		\Return $\mathit{null}$
		\EndFunction
	\end{algorithmic}
\end{algorithm}

The main functionality of the test generator is the generation of diverse functional models.
Pseudo code of the generation process is illustrated in Algorithm~\ref{alg:testgeneration}. It consists of two major functions for 
randomly building a model and for finding a valid model.
The first is a recursive function (Line 1) that starts by selecting an
initial root layer and continues to connect random layers onto it. 
The algorithm increases the probability of stopping based on the $\mathit{level}$ that represents the distance of the current layer to the root layer (Line 2) to prevent too large models.
Many layers can take the output of one layer as input, and some can have inputs from multiple layers, e.g., to perform mathematical operations, like addition, or multiplication of the inputs. Hence, there is a 
variable for the number of inputs, which is set according to the generated layer (Lines 5--7). Finally, the recursion is performed for each input, a connection is formed with the associated functions (by feeding the output of a layer as input to another layer), and the layer is returned.
The model in this case is just the root layer that is connected to other layers. Note that the models produced by this function are most likely invalid, \rev{and for simplicity it produces models in the form of trees. The semantics also supports more complex graphs.}

The second function takes this model as a basis to find a valid model with the help of the semantics (Line 12).
Our semantics can identify invalid models based on layer preconditions, which are defined as part of the corresponding Prolog rules. These preconditions give feedback in the form of a badness value to guide the generation process towards valid models. Given a concrete model and a precondition to be satisfied, the badness value is defined in the same way as fitness in search-based software testing~\cite{DBLP:conf/laser/HarmanMSY10}. \rev{It is a distance metric that increases when the model is further away from becoming valid, i.e., we rank the preconditions based on severity and calculate a value by multiplying a severity factor with the difference of, e.g., an expected argument value to an observed one.}

\begin{figure}[t]
\centering
\includegraphics[width=1\linewidth]{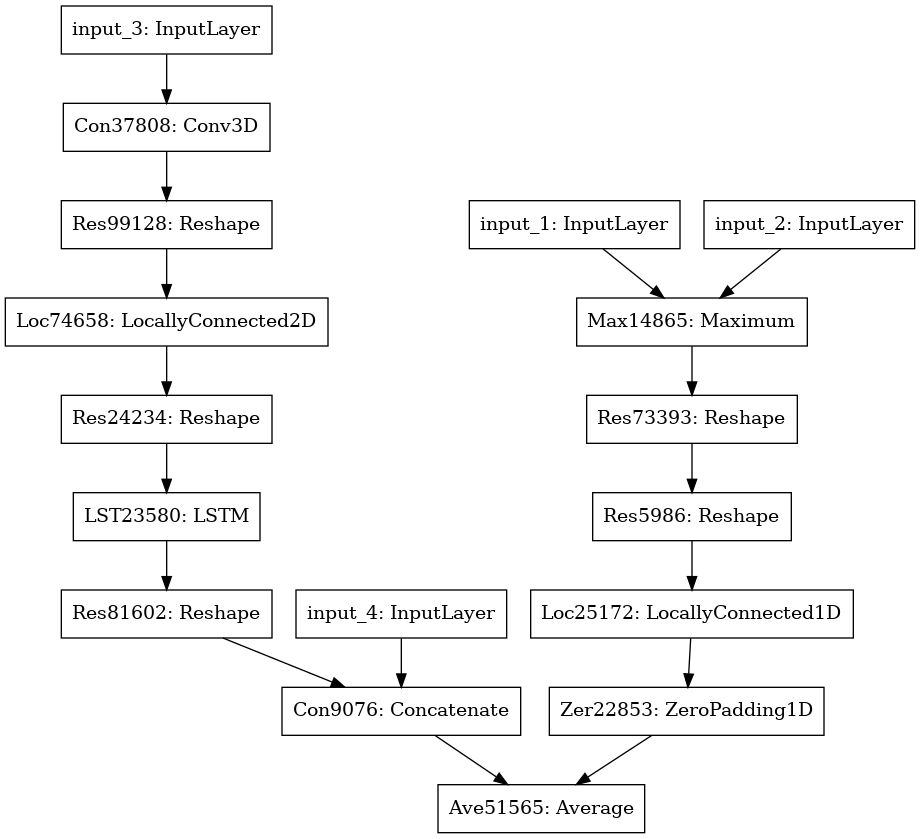}
\caption{Model visualisation in graph form.}
\label{fig:model}
\end{figure}

Line 14 shows that we apply a $\mathit{SemanticHelper}$ to run our semantics and to retrieve a success or an error message. 
If the semantics was executed successful then the model is valid and the 
algorithm stops (Line 16). If an error occurs during the execution, we check if it originates from the same layer as a previous error, and if it is more severe than the last error (based on the badness value), which would indicate that the
last model change did not improve the model. In this case, we restore the last model (Line 18).
Otherwise, we try to adopt the model to gradually improve its validity (Lines 21--32).
This is done by randomly selecting a potential model modification for the layer that causes invalidity, like regenerating arguments, inserting a layer to change the input shapes/dimensions, or replacing the layer with one of its input layers. 
This process is repeated until a valid model is found or a maximum number of tries is reached. 

An example graph model that was generated with our approach is illustrated in Fig.~\ref{fig:model}. The nodes in the model represent layers and the arrows show the data flow. It can be seen that the model includes various layers, like convolutional or reshape, and the output of the model is derived from the root layer, i.e. that average layer at the bottom. The corresponding TensorFlow and Prolog model are available online in our repository \cite{repo}.
\subsection{Model Validation} \label{sec:modelvalidation}
Another application of our semantics is the model validation by identifying bugs in faulty (or invalid) AI models. It can be challenging to create working models, especially since the layers have numerous preconditions that need to be fulfilled when they are used or combined. 
For example, common errors are wrong input shapes and dimensions, or arguments that are invalid for the input data, like a convolution kernel or an axis value that is too large for an input dimensions \cite{DBLP:conf/issta/ZhangCCXZ18,DBLP:conf/icse/HumbatovaJBR0T20}.
AI libraries, like TensorFlow, already produce error message to detect such issues. However, they can be difficult to understand, the messages are not always consistent for the same type of issue, and in rare cases there is no error for a problematic model (see the example below).
Hence, we provide an alternative way to validate models by identifying and reporting issues precisely and consistently with the help of our semantics.
By converting an existing TensorFlow model into our Prolog representation, we can exploit our semantics to check if there are any precondition violations in the model, which would make it invalid.

An example precondition is shown in Listing~\ref{lst:precondition}. It checks if a cropping layer applies too much cropping, which would result in an empty output. There is a predicate \inline|check_empty_cropping| that has the layer input and output as arguments and it is true when the output is not empty. Otherwise, an exception is thrown. First, in Lines~1--2 there is a recursive rule for the inspection of multidimensional output. Next, the rule in Line~3 makes our predicate true, if any numerical value is found in the output. If that is not the case, then we generate an error message Lines~4--7.
\begin{lstlisting}[language=Prolog,  xleftmargin=15pt, float=tp, 
%aboveskip=1pt,belowskip=-5pt,
caption={Prolog code of a layer precondition.}, label={lst:precondition},deletekeywords={is}, morekeywords={lstm_layer,dense_node_comp,depth,add_lists,multiply_list_with,padding1D}]
check_empty_cropping(Is, [O|_]) :-
	check_empty_cropping(Is,O).
check_empty_cropping(_,O) :- number(O).
check_empty_cropping(Is,[]) :-
	writeln("Invalid Model"), S1="Cropping Error, Input Shape ",
	shape(Is,Shape),term_string(Shape,T),string_concat(S1,T,S),
	throw(S).
\end{lstlisting} 

An example TensorFlow model that contains a corresponding cropping issue is shown in Listing~\ref{lst:aibug}. It can be seen that our model consists of four layers, a convolutional layer that is connected with a cropping layer, a maximum pool layer, and a concatenate layer that takes the outputs of the cropping and pool layers.
The same model in Prolog is shown in Listing~\ref{lst:aibugpro}. This model has a slightly different structure as our previous model queries. First, we assign the layers to variables, which we then pass in a list to the \inline|exec_layers| function. This function is a simple wrapper for the execution of a list of layers that helps to identify the name of a layer that violates a precondition.

The model might seem valid at a first glance, 
but investigating the cropping arguments shows that the values are too big.
This causes an empty intermediate output in the cropping layer.
Expectedly, our semantics is able to identify this issue and produces an error.
TensorFlow has no error message \cite{issue6}. The model is just executed, and the concatenation is performed with the empty output, which means that the first branch of the model is completely ignored. Especially in large models, it can be challenging to find such issues since only intermediate results reveal such problems. \rev{The TensorFlow developers confirmed that this should not happen and that an additional check will be added} \cite{issue6}.
\begin{lstlisting}[language=Python, xleftmargin=15pt, float=tp, 
%aboveskip=-6pt,belowskip=-12pt,
caption={Simple example model bug in a TensorFlow.}, label={lst:aibug}]
import tensorflow as tf, numpy as np
from tensorflow.keras import layers, models

in0 = tf.keras.layers.Input(shape=([2, 2]))
in1 = tf.keras.layers.Input(shape=([2, 2]))
Max = layers.MaxPool1D(pool_size=(1), name = 'Max')(in0)
Con = layers.Conv1D(2,(1),padding='valid',name='Con')(in1)
Cro = layers.Cropping1D(cropping=((5, 5)), name = 'Cro')(Con)
Con1 = layers.Concatenate(axis=1, name = 'Con1')([Max,Cro])

model = models.Model(inputs=[in0,in1], outputs=Con1)

w = model.get_layer('Con').get_weights()
w[0] = np.array([[[0.572, 0.621], [0.5388, 0.5741]]])
w[1] = np.array([0, 0])
model.get_layer('Con').set_weights(w)
in0 = tf.constant([[[1.313, 1.02], [1.45, 1.92]]])
in1 = tf.constant([[[0.9421, 0.7879], [0.809, 0.855]]])
print(np.array2string(model.predict([in0,in1],steps=1)))
\end{lstlisting}

Our semantics are not only important for the test generation, but they also enable a reliable way to identify issues in AI models.
Error messages in AI libraries are not always consistent because of multiple (or device-specific) implementations of a layer.
Our preconditions are easy to define and can speed up the identification of issues by highlighting problematic layers and by presenting consistent  messages that could, e.g., support automatic error handling.
\rev{In future work, we aim to provide a more user friendly tool with a graphical user interface in order to facilitate the usage of both our test case generation and model validation approaches in a practical setting, e.g, to produce test models for various purposes and to find issues with existing models.}

\section{Evaluation}\label{sec:evaluation}
We implemented our semantics for almost all TensorFlow layers and developed two example applications, i.e., AI framework testing, and model validation to find  invalid AI models. In this section, we evaluate the effectiveness of the approaches.
We design multiple experiments to answer the following research questions (RQ).

\begin{lstlisting}[language=Prolog,  xleftmargin=15pt, float=tp, 
%aboveskip=-6pt,belowskip=-5pt,
caption={Simple example model bug in a Prolog.}, label={lst:aibugpro},deletekeywords={is}, morekeywords={lstm_layer,dense_node_comp,depth,add_lists,multiply_list_with,padding1D,conv2D_layer,multiply_layer,exec_layers}]
A = max_pool1D_layer([[[1.313,1.02],[1.45,1.92]]],1,1,false,Max),
B = conv1D_layer([[[0.9421,0.7879],[0.809,0.855]]],1,[[[0.572, 0.621],[0.5388, 0.5741]]],[0,0], 1, false, 1, Con),
C = cropping1D_layer(Con, 5, 5, Cro),
D = concatenate_layer([Max,Cro], 1, Con1),
exec_layers([A,B,C,D],["Max","Con","Cro","Con1"],Con1,"Con1")
\end{lstlisting}

\begin{itemize}
	\item  \emph{RQ1: Is Prolog expressive and efficient enough for AI semantics analysis?}
	There are various specification languages that might be suited for developing a semantics of an AI framework. Hence, we want to highlight the advantages of using Prolog. 
	\item \emph{RQ2: Does the semantics allow us to effectively generate valid AI test cases?}
	This is important since one of the applications of our semantics is 
	AI framework testing and for that it is necessary to produce good test models.
	\item \emph{RQ3: Is our test generation method based on the semantics effective for finding AI framework bugs?}
	To further motivate the usage of our testing method, 
	we show what bugs and issues it can reveal.
	\item \emph{RQ4: How well can our semantics report bugs of invalid AI models?} Another application of our 
	semantics is the model validation. We will show the advantages of our method compared with the default bug reporting capabilities of TensorFlow.
\end{itemize}
The experiments were performed on a 7th Gen. Lenovo X1 Carbon ThinkPad with a 8th Gen i7 CPU with four 1.80 GHz cores and 16GB RAM. We mainly tested TensorFlow 2.4, but also discovered issues in version 1.14. For executing the Prolog semantics, we used SWI-Prolog 8.2.1 and the test generator was built in Java 13.0.7.
Below are our answers to the research questions.

\paragraph{RQ1: Is Prolog expressive and efficient enough for AI semantics analysis?}
In order to answer this question, we report our findings on developing our Prolog semantics. From the TensorFlow documentation \cite{layerdocs}, 
we identified 79 unique non-abstract and non-wrapper layers that we wanted to implement. For practical reasons, we could not support all layers, but we were able to develop our specification for 72 layers of various types, like mathematical, normalisation, convolutional, recurrent, pooling, dense, activation, cropping, padding, dropout, and most of these layers support nearly all arguments. \rev{We implemented the core functionality of these layers to be able to perform predictions. Our specification is not concerned with  learning/training algorithms, data preprocessing and model (accuracy) evaluations or optimisations.}
A full list of the layers is available online \cite{repo}.

We did not develop a specification for the remaining 7 layers (ActivityRegularization, AdditiveAttention, Attention, DenseFeatures, Lambda, MultiHeadAttention, StackedRNNCells) since they have too generic or flexible input requirements, like a generic iterable that can contain other layers, or arbitrary functions or queries. 
Moreover, we could not fully support all layer arguments and features, e.g., we implemented the Masking layer as a stand-alone layer, but not the influencing behaviour that it can have on other layers. 
We had to omit some layer arguments since they would have violated our specification requirement of static weights, i.e., there are initialisers, regularisers (for weights, kernels, or biases) that we did not implement since we statically set these attributes. Moreover, we did not implement activation or dropout features within the layers since the same behaviour can be achieved by adding an independent activation, or dropout layer. We left out some arguments, like constraints, data formats, or a trainable option, due to their complexity or because they have an influence outside the scope of our semantics which is not concerned with the training of AI models.
Some layers, like GRU, have an option to choose a specific implementation of the layer. For simplicity, we implemented the default implementations.

\rev{It should be noted that only seven layers had detailed (mathematical) descriptions in the documentation,
44 were explained with examples or had a reference paper, and the remaining 21 layers were under-specified or had just very limited examples.}
From our 72 implemented layers, seven had non-deterministic properties, hence instead of exactly specifying their behaviour, we defined Prolog rules that check if they follow the described properties from the documentations as explained in Sect.~\ref{sec:spec}.
For the remaining 65 layers, we directly specified the functionality, which enables an easy execution and an exact comparison of the prediction results from our semantics to TensorFlow.
The implementation effort of the semantics was about eight work months, which partially also includes the time to acquire the relevant domain knowledge. 

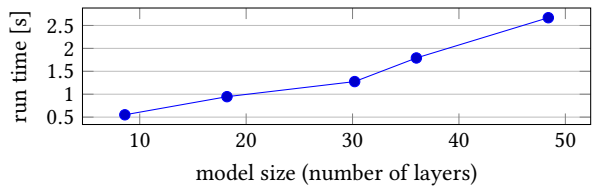
\begin{figure}[!t]
	\centering
	\begin{tikzpicture}[scale=0.98]
	\begin{axis}[,
	try min ticks=5,
	width=1.0\linewidth,
	height=10em,
	xticklabel style={/pgf/number format/precision=6,/pgf/number format/fixed},
	scaled ticks=false,
	ymajorgrids,
	ylabel={run time [s]},
	xlabel={model size (number of layers)},
	]
	\addplot table[x=size,y=duration] {runtime.csv};
	\end{axis}
	\end{tikzpicture}
	\caption{Run time of the semantics for rising model sizes.} 
	\label{fig:runtime}
\end{figure}
We evaluated the execution time of our semantics in order to illustrate the applicability for non-trivial use cases. 
Figure~\ref{fig:runtime} shows the achieved run times for increasing model sizes. We created the models by adopting the level variable of the test generation from Algorithm~\ref{alg:testgeneration}, by setting the input data size to 4KB, and by taking the average over ten models for each data point. 
It can be seen, that for small models with about ten layers and non-trivial inputs the run time of the semantics is only about 0.6s, and even for models with about 50 layers, a prediction can be done in less than 3s. 
Hence, we believe our semantics is fast enough for reasonable models.

A major advantage of using Prolog is its declarative and high-level specification style, which enables a compact and straightforward implementation. 
We only needed about 3200 lines of code for all 72 layers, and on average 44 lines per layer, 
because Prolog has already a lot of built-in functionality, like list operations, and it also facilitates the reuse of code.
In contrast, TensorFlow has a huge code base with about three million lines of code. Hence, we believe our specification provides a good opportunity to AI developers to learn more about the inner workings of layers, and that Prolog is a well suited specification language since it supports a compact and convenient  implementation of most layers. 


\paragraph{RQ2: Does the semantics allow us to effectively generate valid AI test cases?}
In order to evaluate the effectiveness of our test generation approach, we used our test generator to produce 10,000 test models.
The run time of the generation was about 35 hours (on average 12.6s per model), of which most time was spent on gradually finding a valid model with feedback information from the execution of the semantics.
Our generated AI models had an average size of 9.89 layers (excluding input layers), and we used small inputs with a range of one to four values per dimension since bigger sizes vastly increased the overall size of our highly dimensional input data. 

Compared to state-of-the-art approaches for testing AI frameworks, 
our generated models have more diversity in terms of layers and structure.
A related fuzzing method \cite{li2020documentation} only tests single layers and it is unclear how many layers it can test with valid inputs. A differential testing approach \cite{DBLP:conf/sigsoft/WangYCLZ20} can produce valid models with mutation, but they are less diverse since the mutations are only concerned with sequential model aspects, and they can only add 24 types of layers to the mutants. Other related approaches have less variety since they only use existing models or adopt the input data.

Out of the 10,000 test cases 99.1\% were valid, i.e. we were able to produce the same prediction result with our semantics as TensorFlow for the same model and inputs. Seven test cases failed because of a dependency issue of the Masking layer~\cite{masking} that we could not fully support. (Almost all layers work independently, but  
the Masking layer is an exception. It requires all following layers to skip a specific input time step.) 
\rev{Two tests failed due to rare inputs which caused inconsistencies with the TensorFlow input argument checks. We are still investigating this issue because of its rare occurrence.}
Finally, one test failed due to a minor inconstancy in the handling of floating-point values, which only occurred very rarely when a lot of computations with recurrent layers were performed.

\begin{table}[t]
\footnotesize
\caption{Found TensorFlow issues and bugs.}\label{tab:issues}
\begin{tabular}{|l|c|c|c|}
\hline
\textbf{Issue Type} &  \textbf{\begin{tabular}[c]{@{}c@{}}Fixed or \\  Confirmed \end{tabular}} & \textbf{\begin{tabular}[c]{@{}c@{}}Pending or \\  Unconfirmed \end{tabular}} & \textbf{Total} \\ \hline
TensorFlow bugs    &  2  & 1  &  3  \\ \hline
Error message bugs &  4  & 2  &  6 \\ \hline
Documentation bugs &  4  & 1  &  5  \\ \hline
Total              &  10  & 4  & 14  \\ \hline
\end{tabular}
\end{table}
The high percentage of valid AI models that we were able to achieve with our test case generation shows that our method is effective, especially considering the low percentage (25\%) of the related work \cite{li2020documentation}.
Moreover, we were able to reveal various issues and bugs as we will see in the answer to the next research questions.

\paragraph{RQ3: Is our test generation method based on the semantics effective for finding AI framework bugs?}
With our generated test cases, we were able to discover various issues and bugs in TensorFlow. Some of which were actual faults in TensorFlow, a few issues were related to wrong or misleading error messages, and there were some documentation bugs. Table~\ref{tab:issues} shows an overview of the found issues, their type, and if they were confirmed or not. 

The most interesting bug that we found was related to the popular activation function (or layer) called ReLU,
which showed a wrong behaviour for negative thresholds that violated the description of the documentation \cite{issue1}.
\rev{We believe that this bug is the most significant since it is concerned with one of the most common activation functions, and AI models that used this function with a negative threshold (and negative) values might have suffered from inaccuracy or might have produced unexpected prediction results.}
Another bug was regarding the ConvLSTM2D layer, which produced potentially wrong prediction results in rare cases \cite{issue2}, and one was an ignored argument \cite{issue13}.

An issue was regarding an error message that suggested that another number format is required, but when the format was changed then the message stated the original format was needed \cite{issue14}.
There were various similar error message related issues \cite{issue3,issue4,issue5,issue6,issue7}.

A documentation bug was about a missing preconditions description that caused an error when certain arguments were not equal in a SeparableConv 
layer \cite{issue8}. A number of similar issues \cite{issue9,issue10,issue11,issue12} were also related to documentation inconsistencies.

In total, we found 14 issues that we reported to the TensorFlow developers and that helped to improve the framework. Ten issues were confirmed or are already fixed, four are unconfirmed or still pending.
Note that our method focuses on semantic bugs, \rev{like the ReLU bug}~\cite{issue1}, that are harder to find. 
Related work cannot directly be compared due to simpler types of oracles, which we will explain in Sect.~\ref{sec:related}.
Hence, we believe it is reasonable effective, and the fact that we found a major bug in one of the most common activation layers highlights the need for a good executable semantics.

\paragraph{RQ4: How well can our semantics report bugs of invalid AI models?}
In order to evaluate the bug finding or localisation capabilities of our semantics, we generated 100 invalid models by running our test case generation method that gradually finds valid models. 
Retrieving the last model of this process allowed us to have invalid models that
only contain a single bug, which facilitates the manual analysis of the issues.
We evaluated the type of issues that made the models invalid and compared the bug finding and reporting capabilities of our semantics to TensorFlow.

Table~\ref{tab:validation} shows an overview of the types and frequencies of issues, and the number of TensorFlow inconsistencies that we found.
The majority (57) of the models were invalid due to dimension errors caused by layers requiring input data with a maximum, minimum or specific dimension, or when a layer has multiple inputs and their dimensions are not matching.
The next large portion, 29 models were invalid due to inconsistent input data that occurs when a layer takes multiple inputs and there is a requirement that the inputs need to be the same shape or the size of some dimensions must match.
The remaining 14 models were invalid due to argument errors, like a pool or kernel size that is too large, or inconsistencies in weight (or biases) shapes. Most convolutional layers require weight arrays to have a specific shape that can depend on the input size. Hence, a model can, e.g., become invalid when the shape of the weights is inconsistent with the input shape.

\begin{table}[t]
\footnotesize
\caption{Overview of the identified model validation issues.}
\label{tab:validation}
\begin{tabular}{|l|c|c|}
\hline
\textbf{Model Issue} & \textbf{Occurrences} & \textbf{\begin{tabular}[c]{@{}c@{}}TensorFlow \\ Inconsistencies \end{tabular}} \\ \hline
Dimension Error           & 57 & 5  \\ \hline
Inconsistent Input Shapes & 29  & 2  \\ \hline
Argument Error            & 14 & 3 \\ \hline
\end{tabular}
\end{table}

%

Many error messages from TensorFlow were inconsistent  as indicated in the last column of Table~\ref{tab:validation}. Even for simple cases like a wrong input dimension, there are different messages (``ValueError: Input 0 of layer X is incompatible with the layer: expected ndim=5, found ndim=3.'' or  ``ValueError: Shape must be rank 4 but is rank 5 for X.'' or ``ValueError: Inputs should have rank 4. Received input shape \ldots'').
The reason for these differences might be that the error messages are produced independently in different layers. 
In contrast, we can easily reuse a precondition that produces the same message in all layers. This facilitates automatic error processing.

There were cases in which TensorFlow did not produce an error, although there should be one. 
For example, when cropping layers are used with a too large cropping values, then TensorFlow will just produce an empty output \cite{issue6} (more details are in Sect.~\ref{sec:modelvalidation}).
A similar issue was regarding the input shape verification that did not occur with specific prediction methods \cite{issue7}.

Same as TensorFlow, our semantics was able to identify the layers that were causing an invalid model in all cases. In four cases, there were discrepancies regarding the type of error. The reason for that is simple. Some issues can have multiple causes, e.g., when a weight shape is inconsistent with an input, then either of them may be wrong. 

Generally, our produced messages are simpler, consistent, and at a high level. To be fair, TensorFlow's error messages could be at times more detailed (for better debugging), but our focus is on automatic error handling.
Note that there seems to be no other work investigating TensorFlow's error messages for AI models.
Finally, with our approach it is much easier to add preconditions for special use cases or types of layers since the code is much more compact, which makes it easy to find the right place to add a precondition.


\paragraph{Discussion.}
A threat to the validity of our evaluation might be 
that there are limited ways that we can compare with existing works. 
This might have been interesting.
However, there were not many related approaches that could be used for a fair comparison since the closest related work for generating AI models worked with a much weaker oracle and only focused on simpler bugs.

One might argue that our AI framework semantics is limited since it is mostly concerned with the behaviour of layers. AI frameworks have other important components, e.g., it would be interesting to specify the learning algorithms. However, in this work we focused on the static behaviour of layers since we believe it is a good start for more general AI framework semantics, and our specification is valuable 
since it can perform predictions for various complex AI models. 
Moreover, we illustrated how non-deterministic properties can be evaluated with Prolog. 

Another potential threat to the validity of our evaluation could be that the test inputs we used to generate AI models were too small for a realistic study. AI models can handle huge data like large images or videos. Hence, it is true that an evaluation with larger test inputs might be more realistic, but
bugs in the AI frameworks as well as in AI models could be independent of the input size. We believe that our test models with rather small inputs were still reasonable and did not represent a big limitation. Moreover, it is well-known that small test cases can reveal various bugs~\cite{DBLP:conf/issta/JacksonD96}.

One might argue that our model validation approach is not as useful as TensorFlow's bug reporting capabilities. It is true that the produced error messages from TensorFlow are more sophisticated, and they might be better for debugging. However, our goal was to produce simple and consistent error messages that can be used for applications such as automatic error handling or even model repair. Moreover, we wanted to provide an easy way to add new preconditions and reporting capabilities, which can be helpful since not all model issues were identified by TensorFlow. 


\section{Related Work}\label{sec:related}
Executable semantics for programming languages are related to our work because AI libraries have a similar purpose as a compiler. For example, there are various executable semantics for popular programming languages 
that were build with the K framework \cite{DBLP:journals/jlp/RosuS10}. In the same way as our AI framework specification, such semantics can be used as a strong test oracle and facilitate the test generation.

A related approach that also works with AI specifications was presented by Selsam et al.~\cite{DBLP:conf/icml/SelsamLD17}. 
The authors apply an interactive proofing tool and partial synthesis of some system components
to construct an AI system and to prove its correctness. Such a system can also be seen as a formal specification similar to ours since it can be used to detect bugs. However, in this work the authors focus more on an alternative (library) implementation that is correct by design, 
and they only focus on a limited AI system. Moreover, the authors point out that their proof relies on an idealised setting with infinite-precision real numbers that are replaced for a concrete execution, which can be a potential source for bugs.

The closest related work to our testing approach was presented by Wang et al.~\cite{DBLP:conf/sigsoft/WangYCLZ20}. The authors show a differential testing approach called LEMON that uses mutations to produce test cases, and it applies a heuristic strategy to guide the generation in order to amplify  inconsistencies between different DL libraries.
They do this to get rid of noise introduced by non-determinism in DL libraries. With our approach, we can directly test many library components deterministically, which makes the identification of bugs much easier.
Moreover, we do not have to rely on the existence of other libraries. Our semantics can also be used to test newly developed features that are only provided by a single library.

A similar approach called CRADLE was presented by Pham et al.~\cite{DBLP:conf/icse/PhamLQT19}. CRADLE also performs differential testing, and it applies distance metrics to detect inconsistent
outputs. It localises the source of an inconsistency 
by tracking and analysing anomalies in the execution graph. 
Similarly, Srisakaokul et al.~\cite{DBLP:conf/aaai/SrisakaokulWAAX18} show a differential method for supervised learning software and demonstrate it for k-Nearest Neighbour and Naive Bayes implementations.

A related fuzzing method was illustrated by Li \cite{li2020documentation}.
The approach relies on extracting information from AI library documentations to find input requirements of deep learning layers, which are needed to generate test data. In contrast to our work, this method only has very limited capabilities to produce valid tests (about 25\%) and their oracle can only detect documentation bugs or input violations.

Dwarakanath et al.\cite{DBLP:conf/issta/DwarakanathASRB18}
have developed new metamorphic relations for metamorphic testing of 
AI libraries. For example, they perform permutations, rearrangements, or rotations of the input data that should not change the output. Then, they identify bugs based on unexpected changes in the output. They evaluate their method by artificially introducing bugs into image classification systems. 

A similar metamorphic method for validation was presented by Ding et al.~\cite{DBLP:conf/icse/DingKH17}. The approach works with metamorphic relations that, e.g., add (or remove) images or image categories to the training, test or validation data. They also work with image classifiers. Another metamorphic method that permutates rows and columns, shuffles and renames input features in order to evaluate the balancedness of machine learning classifiers was introduced by Sharma and Wehrheim~\cite{DBLP:conf/icst/SharmaW19}.
In contrast to these approaches, our semantics will work with much more AI applications, and its strong oracle supports the detection of a wider range of bugs.

Related to our model validation method are some 
debugging approaches \cite{DBLP:journals/tvcg/StrobeltGBPPR19,DBLP:conf/chi/SchoopHH21,DBLP:journals/tvcg/HohmanKPC19} for AI models that offer features like visualisation or auditing, to find issues. However, such approaches are usually limited to specific types of models, e.g., classifiers, and their focus is not on introducing custom validation options.

\section{Conclusion} \label{sec:conclusion}
We have introduced a novel executable semantics for AI frameworks and illustrated two of its applications.
Our semantics is implemented in Prolog and it focuses on the deterministic behaviour of deep learning layers. With Prolog, we could develop 
almost all layers of TensorFlow in a convenient and compact specification style.

One major application of our semantics is the automatic generation of test AI models. We applied a fuzzing approach that incorporates feedback from the semantics to gradually find valid models. 
The approach was able to consistently produce valid models in 99\% of the cases, which is much higher than related approaches.
Moreover, we evaluated the effectiveness of this method for finding AI framework bugs, and the results were encouraging.
We discovered various issues and bugs in TensorFlow and could thereby help to improve this AI framework. Most notably, we even found a bug in the well-established ReLU function.

Another application that we presented is the validation of deep learning models. Our semantics is build with various preconditions that can find bugs in invalid AI models. The evaluation showed that we can produce consistent error messages, and we were able to identify issues that could not be discovered by TensorFlow.

We believe that these two approaches highlight the usefulness and generality of our semantics, which will also enable further applications.
In the future, we aim to explore different test generation methods and AI model synthesis.

\subsubsection*{Acknowledgments.}
This project is supported by the National Research Foundation, Singapore and National University of Singapore through its National Satellite of Excellence in Trustworthy Software Systems (NSOE-TSS) office under the Trustworthy Software Systems -- Core Technologies Grant (TSSCTG) award no. NSOE-TSS2019-03. 

Any opinions, findings and conclusions or recommendations expressed in this material are those of the author(s) and do not reflect the views of National Research Foundation, Singapore and National University of Singapore (including its National Satellite of Excellence in Trustworthy Software Systems (NSOE-TSS) office).

\rev{Additionally, we would like to thank Mengdi Zhang and Ayesha Sadiq for their help with some experiments of this work.}

\balance
\bibliographystyle{ACM-Reference-Format}
\bibliography{main}


\begin{thebibliography}{45}


\ifx \showCODEN    \undefined \def \showCODEN     #1{\unskip}     \fi
\ifx \showDOI      \undefined \def \showDOI       #1{#1}\fi
\ifx \showISBNx    \undefined \def \showISBNx     #1{\unskip}     \fi
\ifx \showISBNxiii \undefined \def \showISBNxiii  #1{\unskip}     \fi
\ifx \showISSN     \undefined \def \showISSN      #1{\unskip}     \fi
\ifx \showLCCN     \undefined \def \showLCCN      #1{\unskip}     \fi
\ifx \shownote     \undefined \def \shownote      #1{#1}          \fi
\ifx \showarticletitle \undefined \def \showarticletitle #1{#1}   \fi
\ifx \showURL      \undefined \def \showURL       {\relax}        \fi
\providecommand\bibfield[2]{#2}
\providecommand\bibinfo[2]{#2}
\providecommand\natexlab[1]{#1}
\providecommand\showeprint[2][]{arXiv:#2}

\bibitem[\protect\citeauthoryear{??}{iss}{2020}]%
        {issue9}
 \bibinfo{year}{2020}\natexlab{}.
\newblock \bibinfo{title}{Wrong default value in GRU layer documentaion}.
\newblock
\newblock
\urldef\tempurl%
\url{https://github.com/tensorflow/tensorflow/issues/45705}
\showURL{%
\tempurl}


\bibitem[\protect\citeauthoryear{??}{con}{2021}]%
        {conv}
 \bibinfo{year}{2021}\natexlab{}.
\newblock \bibinfo{title}{1D Convolution Layer}.
\newblock
\newblock
\urldef\tempurl%
\url{https://www.tensorflow.org/api_docs/python/tf/keras/layers/Conv1D}
\showURL{%
\tempurl}


\bibitem[\protect\citeauthoryear{??}{iss}{2021a}]%
        {issue4}
 \bibinfo{year}{2021}\natexlab{a}.
\newblock \bibinfo{title}{AveragePooling3D does not support float64 and
  produces confusing error message}.
\newblock
\newblock
\urldef\tempurl%
\url{https://github.com/tensorflow/tensorflow/issues/48644}
\showURL{%
\tempurl}


\bibitem[\protect\citeauthoryear{??}{iss}{2021b}]%
        {issue2}
 \bibinfo{year}{2021}\natexlab{b}.
\newblock \bibinfo{title}{ConvLSTM2D layer wrong computation}.
\newblock
\newblock
\urldef\tempurl%
\url{https://github.com/keras-team/keras/issues/15224}
\showURL{%
\tempurl}


\bibitem[\protect\citeauthoryear{??}{iss}{2021c}]%
        {issue6}
 \bibinfo{year}{2021}\natexlab{c}.
\newblock \bibinfo{title}{cropping layer additional error message}.
\newblock
\newblock
\urldef\tempurl%
\url{https://github.com/tensorflow/tensorflow/issues/50612}
\showURL{%
\tempurl}


\bibitem[\protect\citeauthoryear{??}{den}{2021}]%
        {dense}
 \bibinfo{year}{2021}\natexlab{}.
\newblock \bibinfo{title}{Dense Layer}.
\newblock
\newblock
\urldef\tempurl%
\url{https://www.tensorflow.org/api_docs/python/tf/keras/layers/Dense}
\showURL{%
\tempurl}


\bibitem[\protect\citeauthoryear{??}{iss}{2021d}]%
        {issue10}
 \bibinfo{year}{2021}\natexlab{d}.
\newblock \bibinfo{title}{Dot layer incomplete description}.
\newblock
\newblock
\urldef\tempurl%
\url{https://github.com/tensorflow/tensorflow/issues/45706}
\showURL{%
\tempurl}


\bibitem[\protect\citeauthoryear{??}{dro}{2021}]%
        {dropout}
 \bibinfo{year}{2021}\natexlab{}.
\newblock \bibinfo{title}{Dropout Layer}.
\newblock
\newblock
\urldef\tempurl%
\url{https://www.tensorflow.org/api_docs/python/tf/keras/layers/Dropout}
\showURL{%
\tempurl}


\bibitem[\protect\citeauthoryear{??}{iss}{2021e}]%
        {issue7}
 \bibinfo{year}{2021}\natexlab{e}.
\newblock \bibinfo{title}{error reporting model(x) vs model.predict(x)}.
\newblock
\newblock
\urldef\tempurl%
\url{https://github.com/tensorflow/tensorflow/issues/50618}
\showURL{%
\tempurl}


\bibitem[\protect\citeauthoryear{??}{rep}{2021}]%
        {repo}
 \bibinfo{year}{2021}\natexlab{}.
\newblock \bibinfo{title}{ExAIS: Executable AI Semantics Repository}.
\newblock
\newblock
\urldef\tempurl%
\url{https://github.com/rschumi0/ExAIS}
\showURL{%
\tempurl}


\bibitem[\protect\citeauthoryear{??}{iss}{2021f}]%
        {issue13}
 \bibinfo{year}{2021}\natexlab{f}.
\newblock \bibinfo{title}{InputSpec argument ignored}.
\newblock
\newblock
\urldef\tempurl%
\url{https://github.com/keras-team/keras/issues/15225}
\showURL{%
\tempurl}


\bibitem[\protect\citeauthoryear{??}{iss}{2021g}]%
        {issue14}
 \bibinfo{year}{2021}\natexlab{g}.
\newblock \bibinfo{title}{InputSpec missing float64 support and wrong error
  message}.
\newblock
\newblock
\urldef\tempurl%
\url{https://github.com/keras-team/keras/issues/15226}
\showURL{%
\tempurl}


\bibitem[\protect\citeauthoryear{??}{iss}{2021h}]%
        {issue12}
 \bibinfo{year}{2021}\natexlab{h}.
\newblock \bibinfo{title}{layer order in functional API graph models}.
\newblock
\newblock
\urldef\tempurl%
\url{https://github.com/tensorflow/tensorflow/issues/50306}
\showURL{%
\tempurl}


\bibitem[\protect\citeauthoryear{??}{mas}{2021}]%
        {masking}
 \bibinfo{year}{2021}\natexlab{}.
\newblock \bibinfo{title}{Masking Layer}.
\newblock
\newblock
\urldef\tempurl%
\url{https://www.tensorflow.org/api_docs/python/tf/keras/layers/Masking}
\showURL{%
\tempurl}


\bibitem[\protect\citeauthoryear{??}{iss}{2021i}]%
        {issue1}
 \bibinfo{year}{2021}\natexlab{i}.
\newblock \bibinfo{title}{ReLU layer wrong result with negative threshold}.
\newblock
\newblock
\urldef\tempurl%
\url{https://github.com/tensorflow/tensorflow/issues/48646}
\showURL{%
\tempurl}


\bibitem[\protect\citeauthoryear{??}{iss}{2021j}]%
        {issue8}
 \bibinfo{year}{2021}\natexlab{j}.
\newblock \bibinfo{title}{SeparableConv documention missing argument
  constraint}.
\newblock
\newblock
\urldef\tempurl%
\url{https://github.com/tensorflow/tensorflow/issues/45259}
\showURL{%
\tempurl}


\bibitem[\protect\citeauthoryear{??}{iss}{2021k}]%
        {issue3}
 \bibinfo{year}{2021}\natexlab{k}.
\newblock \bibinfo{title}{Softmax layer unexpected and confusing error
  message}.
\newblock
\newblock
\urldef\tempurl%
\url{https://github.com/tensorflow/tensorflow/issues/50467}
\showURL{%
\tempurl}


\bibitem[\protect\citeauthoryear{??}{iss}{2021l}]%
        {issue11}
 \bibinfo{year}{2021}\natexlab{l}.
\newblock \bibinfo{title}{Softmax layer unexpected behaviour for axis=0}.
\newblock
\newblock
\urldef\tempurl%
\url{https://github.com/tensorflow/tensorflow/issues/48647}
\showURL{%
\tempurl}


\bibitem[\protect\citeauthoryear{??}{lay}{2021}]%
        {layerdocs}
 \bibinfo{year}{2021}\natexlab{}.
\newblock \bibinfo{title}{TensorFlow layer documentation}.
\newblock
\newblock
\urldef\tempurl%
\url{https://www.tensorflow.org/api_docs/python/tf/keras/layers}
\showURL{%
\tempurl}


\bibitem[\protect\citeauthoryear{??}{iss}{2021m}]%
        {issue5}
 \bibinfo{year}{2021}\natexlab{m}.
\newblock \bibinfo{title}{Wrong error message for DepthwiseConv2D}.
\newblock
\newblock
\urldef\tempurl%
\url{https://github.com/tensorflow/tensorflow/issues/45703}
\showURL{%
\tempurl}


\bibitem[\protect\citeauthoryear{Bramer}{Bramer}{2013}]%
        {DBLP:books/sp/Bramer13}
\bibfield{author}{\bibinfo{person}{Max Bramer}.}
  \bibinfo{year}{2013}\natexlab{}.
\newblock \bibinfo{booktitle}{\emph{Logic Programming with Prolog}}.
\newblock \bibinfo{publisher}{Springer}.
\newblock
\showISBNx{978-1-4471-5486-0}
\urldef\tempurl%
\url{https://doi.org/10.1007/978-1-4471-5487-7}
\showDOI{\tempurl}


\bibitem[\protect\citeauthoryear{Ding, Kang, and Hu}{Ding
  et~al\mbox{.}}{2017}]%
        {DBLP:conf/icse/DingKH17}
\bibfield{author}{\bibinfo{person}{Junhua Ding}, \bibinfo{person}{Xiaojun
  Kang}, {and} \bibinfo{person}{Xin{-}Hua Hu}.}
  \bibinfo{year}{2017}\natexlab{}.
\newblock \showarticletitle{Validating a Deep Learning Framework by Metamorphic
  Testing}. In \bibinfo{booktitle}{\emph{2nd {IEEE/ACM} International Workshop
  on Metamorphic Testing, MET@ICSE 2017, Buenos Aires, Argentina, May 22,
  2017}}. \bibinfo{publisher}{{IEEE} Computer Society},
  \bibinfo{pages}{28--34}.
\newblock
\urldef\tempurl%
\url{https://doi.org/10.1109/MET.2017.2}
\showDOI{\tempurl}


\bibitem[\protect\citeauthoryear{Dwarakanath, Ahuja, Sikand, Rao, Bose, Dubash,
  and Podder}{Dwarakanath et~al\mbox{.}}{2018}]%
        {DBLP:conf/issta/DwarakanathASRB18}
\bibfield{author}{\bibinfo{person}{Anurag Dwarakanath}, \bibinfo{person}{Manish
  Ahuja}, \bibinfo{person}{Samarth Sikand}, \bibinfo{person}{Raghotham~M. Rao},
  \bibinfo{person}{R.~P. Jagadeesh~Chandra Bose}, \bibinfo{person}{Neville
  Dubash}, {and} \bibinfo{person}{Sanjay Podder}.}
  \bibinfo{year}{2018}\natexlab{}.
\newblock \showarticletitle{Identifying implementation bugs in machine learning
  based image classifiers using metamorphic testing}. In
  \bibinfo{booktitle}{\emph{Proceedings of the 27th {ACM} {SIGSOFT}
  International Symposium on Software Testing and Analysis, {ISSTA} 2018,
  Amsterdam, The Netherlands, July 16-21, 2018}},
  \bibfield{editor}{\bibinfo{person}{Frank Tip} {and} \bibinfo{person}{Eric
  Bodden}} (Eds.). \bibinfo{publisher}{{ACM}}, \bibinfo{pages}{118--128}.
\newblock
\urldef\tempurl%
\url{https://doi.org/10.1145/3213846.3213858}
\showDOI{\tempurl}


\bibitem[\protect\citeauthoryear{Fehlmann}{Fehlmann}{2020}]%
        {fehlmann2020autonomous}
\bibfield{author}{\bibinfo{person}{Thomas~Michael Fehlmann}.}
  \bibinfo{year}{2020}\natexlab{}.
\newblock \bibinfo{booktitle}{\emph{Autonomous Real-Time Testing: Testing
  Artificial Intelligence and Other Complex Systems}}.
\newblock \bibinfo{publisher}{Logos Verlag Berlin GmbH}.
\newblock


\bibitem[\protect\citeauthoryear{Harman, McMinn, {Teixeira de Souza}, and
  Yoo}{Harman et~al\mbox{.}}{2010}]%
        {DBLP:conf/laser/HarmanMSY10}
\bibfield{author}{\bibinfo{person}{Mark Harman}, \bibinfo{person}{Phil McMinn},
  \bibinfo{person}{Jerffeson {Teixeira de Souza}}, {and} \bibinfo{person}{Shin
  Yoo}.} \bibinfo{year}{2010}\natexlab{}.
\newblock \showarticletitle{Search Based Software Engineering: Techniques,
  Taxonomy, Tutorial}. In \bibinfo{booktitle}{\emph{Empirical Software
  Engineering and Verification - International Summer Schools, {LASER}
  2008-2010, Elba Island, Italy, Revised Tutorial Lectures}}
  \emph{(\bibinfo{series}{Lecture Notes in Computer Science})},
  \bibfield{editor}{\bibinfo{person}{Bertrand Meyer} {and}
  \bibinfo{person}{Martin Nordio}} (Eds.), Vol.~\bibinfo{volume}{7007}.
  \bibinfo{publisher}{Springer}, \bibinfo{pages}{1--59}.
\newblock
\urldef\tempurl%
\url{https://doi.org/10.1007/978-3-642-25231-0_1}
\showDOI{\tempurl}


\bibitem[\protect\citeauthoryear{Hohman, Kahng, Pienta, and Chau}{Hohman
  et~al\mbox{.}}{2019}]%
        {DBLP:journals/tvcg/HohmanKPC19}
\bibfield{author}{\bibinfo{person}{Fred Hohman}, \bibinfo{person}{Minsuk
  Kahng}, \bibinfo{person}{Robert Pienta}, {and} \bibinfo{person}{Duen~Horng
  Chau}.} \bibinfo{year}{2019}\natexlab{}.
\newblock \showarticletitle{Visual Analytics in Deep Learning: An Interrogative
  Survey for the Next Frontiers}.
\newblock \bibinfo{journal}{\emph{{IEEE} Trans. Vis. Comput. Graph.}}
  \bibinfo{volume}{25}, \bibinfo{number}{8} (\bibinfo{year}{2019}),
  \bibinfo{pages}{2674--2693}.
\newblock
\urldef\tempurl%
\url{https://doi.org/10.1109/TVCG.2018.2843369}
\showDOI{\tempurl}


\bibitem[\protect\citeauthoryear{Humbatova, Jahangirova, Bavota, Riccio,
  Stocco, and Tonella}{Humbatova et~al\mbox{.}}{2020}]%
        {DBLP:conf/icse/HumbatovaJBR0T20}
\bibfield{author}{\bibinfo{person}{Nargiz Humbatova}, \bibinfo{person}{Gunel
  Jahangirova}, \bibinfo{person}{Gabriele Bavota}, \bibinfo{person}{Vincenzo
  Riccio}, \bibinfo{person}{Andrea Stocco}, {and} \bibinfo{person}{Paolo
  Tonella}.} \bibinfo{year}{2020}\natexlab{}.
\newblock \showarticletitle{Taxonomy of real faults in deep learning systems}.
  In \bibinfo{booktitle}{\emph{{ICSE} '20: 42nd International Conference on
  Software Engineering, Seoul, South Korea, 27 June - 19 July, 2020}},
  \bibfield{editor}{\bibinfo{person}{Gregg Rothermel} {and}
  \bibinfo{person}{Doo{-}Hwan Bae}} (Eds.). \bibinfo{publisher}{{ACM}},
  \bibinfo{pages}{1110--1121}.
\newblock
\urldef\tempurl%
\url{https://doi.org/10.1145/3377811.3380395}
\showDOI{\tempurl}


\bibitem[\protect\citeauthoryear{Jackson and Damon}{Jackson and Damon}{1996}]%
        {DBLP:conf/issta/JacksonD96}
\bibfield{author}{\bibinfo{person}{Daniel Jackson} {and} \bibinfo{person}{Craig
  Damon}.} \bibinfo{year}{1996}\natexlab{}.
\newblock \showarticletitle{Elements of Style: Analyzing a Software Design
  Feature with a Counterexample Detector}. In
  \bibinfo{booktitle}{\emph{Proceedings of the 1996 International Symposium on
  Software Testing and Analysis, {ISSTA} 1996, San Diego, CA, USA, January
  8-10, 1996}}. \bibinfo{publisher}{{ACM}}, \bibinfo{pages}{239--249}.
\newblock
\urldef\tempurl%
\url{https://doi.org/10.1145/229000.226322}
\showDOI{\tempurl}


\bibitem[\protect\citeauthoryear{Li}{Li}{2020}]%
        {li2020documentation}
\bibfield{author}{\bibinfo{person}{Yitong Li}.}
  \bibinfo{year}{2020}\natexlab{}.
\newblock \emph{\bibinfo{title}{Documentation-Guided Fuzzing for Testing Deep
  Learning API Functions}}.
\newblock \bibinfo{thesistype}{Master's\ thesis}. \bibinfo{school}{University
  of Waterloo}.
\newblock


\bibitem[\protect\citeauthoryear{Ma, Zhang, Xue, Li, Liu, Zhao, and Wang}{Ma
  et~al\mbox{.}}{2018}]%
        {DBLP:journals/corr/abs-1806-07723}
\bibfield{author}{\bibinfo{person}{Lei Ma}, \bibinfo{person}{Fuyuan Zhang},
  \bibinfo{person}{Minhui Xue}, \bibinfo{person}{Bo Li}, \bibinfo{person}{Yang
  Liu}, \bibinfo{person}{Jianjun Zhao}, {and} \bibinfo{person}{Yadong Wang}.}
  \bibinfo{year}{2018}\natexlab{}.
\newblock \showarticletitle{Combinatorial Testing for Deep Learning Systems}.
\newblock \bibinfo{journal}{\emph{CoRR}}  \bibinfo{volume}{abs/1806.07723}
  (\bibinfo{year}{2018}).
\newblock
\urldef\tempurl%
\url{http://arxiv.org/abs/1806.07723}
\showURL{%
\tempurl}


\bibitem[\protect\citeauthoryear{Nejadgholi and Yang}{Nejadgholi and
  Yang}{2019}]%
        {DBLP:conf/kbse/NejadgholiY19}
\bibfield{author}{\bibinfo{person}{Mahdi Nejadgholi} {and}
  \bibinfo{person}{Jinqiu Yang}.} \bibinfo{year}{2019}\natexlab{}.
\newblock \showarticletitle{A Study of Oracle Approximations in Testing Deep
  Learning Libraries}. In \bibinfo{booktitle}{\emph{34th {IEEE/ACM}
  International Conference on Automated Software Engineering, {ASE} 2019, San
  Diego, CA, USA, November 11-15, 2019}}. \bibinfo{publisher}{{IEEE}},
  \bibinfo{pages}{785--796}.
\newblock
\urldef\tempurl%
\url{https://doi.org/10.1109/ASE.2019.00078}
\showDOI{\tempurl}


\bibitem[\protect\citeauthoryear{Nilsson and Ma{\l}uszy{\'n}ski}{Nilsson and
  Ma{\l}uszy{\'n}ski}{1990}]%
        {nilsson1990logic}
\bibfield{author}{\bibinfo{person}{Ulf Nilsson} {and} \bibinfo{person}{Jan
  Ma{\l}uszy{\'n}ski}.} \bibinfo{year}{1990}\natexlab{}.
\newblock \bibinfo{booktitle}{\emph{Logic, programming and {Prolog}}}.
\newblock \bibinfo{publisher}{Wiley Chichester}.
\newblock


\bibitem[\protect\citeauthoryear{Pham, Lutellier, Qi, and Tan}{Pham
  et~al\mbox{.}}{2019}]%
        {DBLP:conf/icse/PhamLQT19}
\bibfield{author}{\bibinfo{person}{Hung~Viet Pham}, \bibinfo{person}{Thibaud
  Lutellier}, \bibinfo{person}{Weizhen Qi}, {and} \bibinfo{person}{Lin Tan}.}
  \bibinfo{year}{2019}\natexlab{}.
\newblock \showarticletitle{{CRADLE:} cross-backend validation to detect and
  localize bugs in deep learning libraries}. In
  \bibinfo{booktitle}{\emph{Proceedings of the 41st International Conference on
  Software Engineering, {ICSE} 2019, Montreal, QC, Canada, May 25-31, 2019}},
  \bibfield{editor}{\bibinfo{person}{Joanne~M. Atlee}, \bibinfo{person}{Tevfik
  Bultan}, {and} \bibinfo{person}{Jon Whittle}} (Eds.).
  \bibinfo{publisher}{{IEEE} / {ACM}}, \bibinfo{pages}{1027--1038}.
\newblock
\urldef\tempurl%
\url{https://doi.org/10.1109/ICSE.2019.00107}
\showDOI{\tempurl}


\bibitem[\protect\citeauthoryear{Pouyanfar, Sadiq, Yan, Tian, Tao, Reyes, Shyu,
  Chen, and Iyengar}{Pouyanfar et~al\mbox{.}}{2019}]%
        {DBLP:journals/csur/PouyanfarSYTTRS19}
\bibfield{author}{\bibinfo{person}{Samira Pouyanfar}, \bibinfo{person}{Saad
  Sadiq}, \bibinfo{person}{Yilin Yan}, \bibinfo{person}{Haiman Tian},
  \bibinfo{person}{Yudong Tao}, \bibinfo{person}{Maria E.~Presa Reyes},
  \bibinfo{person}{Mei{-}Ling Shyu}, \bibinfo{person}{Shu{-}Ching Chen}, {and}
  \bibinfo{person}{S.~S. Iyengar}.} \bibinfo{year}{2019}\natexlab{}.
\newblock \showarticletitle{A Survey on Deep Learning: Algorithms, Techniques,
  and Applications}.
\newblock \bibinfo{journal}{\emph{{ACM} Comput. Surv.}} \bibinfo{volume}{51},
  \bibinfo{number}{5} (\bibinfo{year}{2019}), \bibinfo{pages}{92:1--92:36}.
\newblock
\urldef\tempurl%
\url{https://doi.org/10.1145/3234150}
\showDOI{\tempurl}


\bibitem[\protect\citeauthoryear{Rosu and Serbanuta}{Rosu and
  Serbanuta}{2010}]%
        {DBLP:journals/jlp/RosuS10}
\bibfield{author}{\bibinfo{person}{Grigore Rosu} {and}
  \bibinfo{person}{Traian{-}Florin Serbanuta}.}
  \bibinfo{year}{2010}\natexlab{}.
\newblock \showarticletitle{An overview of the {K} semantic framework}.
\newblock \bibinfo{journal}{\emph{J. Log. Algebraic Methods Program.}}
  \bibinfo{volume}{79}, \bibinfo{number}{6} (\bibinfo{year}{2010}),
  \bibinfo{pages}{397--434}.
\newblock
\urldef\tempurl%
\url{https://doi.org/10.1016/j.jlap.2010.03.012}
\showDOI{\tempurl}


\bibitem[\protect\citeauthoryear{Schmidhuber}{Schmidhuber}{2015}]%
        {DBLP:journals/nn/Schmidhuber15}
\bibfield{author}{\bibinfo{person}{J{\"{u}}rgen Schmidhuber}.}
  \bibinfo{year}{2015}\natexlab{}.
\newblock \showarticletitle{Deep learning in neural networks: An overview}.
\newblock \bibinfo{journal}{\emph{Neural Networks}}  \bibinfo{volume}{61}
  (\bibinfo{year}{2015}), \bibinfo{pages}{85--117}.
\newblock
\urldef\tempurl%
\url{https://doi.org/10.1016/j.neunet.2014.09.003}
\showDOI{\tempurl}


\bibitem[\protect\citeauthoryear{Schoop, Huang, and Hartmann}{Schoop
  et~al\mbox{.}}{2021}]%
        {DBLP:conf/chi/SchoopHH21}
\bibfield{author}{\bibinfo{person}{Eldon Schoop}, \bibinfo{person}{Forrest
  Huang}, {and} \bibinfo{person}{Bjoern Hartmann}.}
  \bibinfo{year}{2021}\natexlab{}.
\newblock \showarticletitle{{UMLAUT:} Debugging Deep Learning Programs using
  Program Structure and Model Behavior}. In \bibinfo{booktitle}{\emph{{CHI}
  '21: {CHI} Conference on Human Factors in Computing Systems, Virtual Event /
  Yokohama, Japan, May 8-13, 2021}},
  \bibfield{editor}{\bibinfo{person}{Yoshifumi Kitamura},
  \bibinfo{person}{Aaron Quigley}, \bibinfo{person}{Katherine Isbister},
  \bibinfo{person}{Takeo Igarashi}, \bibinfo{person}{Pernille Bj{\o}rn}, {and}
  \bibinfo{person}{Steven~Mark Drucker}} (Eds.). \bibinfo{publisher}{{ACM}},
  \bibinfo{pages}{310:1--310:16}.
\newblock
\urldef\tempurl%
\url{https://doi.org/10.1145/3411764.3445538}
\showDOI{\tempurl}


\bibitem[\protect\citeauthoryear{Selsam, Liang, and Dill}{Selsam
  et~al\mbox{.}}{2017}]%
        {DBLP:conf/icml/SelsamLD17}
\bibfield{author}{\bibinfo{person}{Daniel Selsam}, \bibinfo{person}{Percy
  Liang}, {and} \bibinfo{person}{David~L. Dill}.}
  \bibinfo{year}{2017}\natexlab{}.
\newblock \showarticletitle{Developing Bug-Free Machine Learning Systems With
  Formal Mathematics}. In \bibinfo{booktitle}{\emph{Proceedings of the 34th
  International Conference on Machine Learning, {ICML} 2017, Sydney, NSW,
  Australia, 6-11 August 2017}} \emph{(\bibinfo{series}{Proceedings of Machine
  Learning Research})}, \bibfield{editor}{\bibinfo{person}{Doina Precup} {and}
  \bibinfo{person}{Yee~Whye Teh}} (Eds.), Vol.~\bibinfo{volume}{70}.
  \bibinfo{publisher}{{PMLR}}, \bibinfo{pages}{3047--3056}.
\newblock
\urldef\tempurl%
\url{http://proceedings.mlr.press/v70/selsam17a.html}
\showURL{%
\tempurl}


\bibitem[\protect\citeauthoryear{Serbanuta, Rosu, and Meseguer}{Serbanuta
  et~al\mbox{.}}{2009}]%
        {DBLP:journals/iandc/SerbanutaRM09}
\bibfield{author}{\bibinfo{person}{Traian{-}Florin Serbanuta},
  \bibinfo{person}{Grigore Rosu}, {and} \bibinfo{person}{Jos{\'{e}} Meseguer}.}
  \bibinfo{year}{2009}\natexlab{}.
\newblock \showarticletitle{A rewriting logic approach to operational
  semantics}.
\newblock \bibinfo{journal}{\emph{Inf. Comput.}} \bibinfo{volume}{207},
  \bibinfo{number}{2} (\bibinfo{year}{2009}), \bibinfo{pages}{305--340}.
\newblock
\urldef\tempurl%
\url{https://doi.org/10.1016/j.ic.2008.03.026}
\showDOI{\tempurl}


\bibitem[\protect\citeauthoryear{Sharma and Wehrheim}{Sharma and
  Wehrheim}{2019}]%
        {DBLP:conf/icst/SharmaW19}
\bibfield{author}{\bibinfo{person}{Arnab Sharma} {and} \bibinfo{person}{Heike
  Wehrheim}.} \bibinfo{year}{2019}\natexlab{}.
\newblock \showarticletitle{Testing Machine Learning Algorithms for Balanced
  Data Usage}. In \bibinfo{booktitle}{\emph{12th {IEEE} Conference on Software
  Testing, Validation and Verification, {ICST} 2019, Xi'an, China, April 22-27,
  2019}}. \bibinfo{publisher}{{IEEE}}, \bibinfo{pages}{125--135}.
\newblock
\urldef\tempurl%
\url{https://doi.org/10.1109/ICST.2019.00022}
\showDOI{\tempurl}


\bibitem[\protect\citeauthoryear{Srisakaokul, Wu, Astorga, Alebiosu, and
  Xie}{Srisakaokul et~al\mbox{.}}{2018}]%
        {DBLP:conf/aaai/SrisakaokulWAAX18}
\bibfield{author}{\bibinfo{person}{Siwakorn Srisakaokul},
  \bibinfo{person}{Zhengkai Wu}, \bibinfo{person}{Angello Astorga},
  \bibinfo{person}{Oreoluwa Alebiosu}, {and} \bibinfo{person}{Tao Xie}.}
  \bibinfo{year}{2018}\natexlab{}.
\newblock \showarticletitle{Multiple-Implementation Testing of Supervised
  Learning Software}. In \bibinfo{booktitle}{\emph{The Workshops of the The
  Thirty-Second {AAAI} Conference on Artificial Intelligence, New Orleans,
  Louisiana, USA, February 2-7, 2018}} \emph{(\bibinfo{series}{{AAAI}
  Workshops})}, Vol.~\bibinfo{volume}{{WS-18}}. \bibinfo{publisher}{{AAAI}
  Press}, \bibinfo{pages}{384--391}.
\newblock
\urldef\tempurl%
\url{https://aaai.org/ocs/index.php/WS/AAAIW18/paper/view/17301}
\showURL{%
\tempurl}


\bibitem[\protect\citeauthoryear{Strobelt, Gehrmann, Behrisch, Perer, Pfister,
  and Rush}{Strobelt et~al\mbox{.}}{2019}]%
        {DBLP:journals/tvcg/StrobeltGBPPR19}
\bibfield{author}{\bibinfo{person}{Hendrik Strobelt},
  \bibinfo{person}{Sebastian Gehrmann}, \bibinfo{person}{Michael Behrisch},
  \bibinfo{person}{Adam Perer}, \bibinfo{person}{Hanspeter Pfister}, {and}
  \bibinfo{person}{Alexander~M. Rush}.} \bibinfo{year}{2019}\natexlab{}.
\newblock \showarticletitle{Seq2seq-Vis: {A} Visual Debugging Tool for
  Sequence-to-Sequence Models}.
\newblock \bibinfo{journal}{\emph{{IEEE} Trans. Vis. Comput. Graph.}}
  \bibinfo{volume}{25}, \bibinfo{number}{1} (\bibinfo{year}{2019}),
  \bibinfo{pages}{353--363}.
\newblock
\urldef\tempurl%
\url{https://doi.org/10.1109/TVCG.2018.2865044}
\showDOI{\tempurl}


\bibitem[\protect\citeauthoryear{Wang, Yan, Chen, Liu, and Zhang}{Wang
  et~al\mbox{.}}{2020}]%
        {DBLP:conf/sigsoft/WangYCLZ20}
\bibfield{author}{\bibinfo{person}{Zan Wang}, \bibinfo{person}{Ming Yan},
  \bibinfo{person}{Junjie Chen}, \bibinfo{person}{Shuang Liu}, {and}
  \bibinfo{person}{Dongdi Zhang}.} \bibinfo{year}{2020}\natexlab{}.
\newblock \showarticletitle{Deep learning library testing via effective model
  generation}. In \bibinfo{booktitle}{\emph{{ESEC/FSE} '20: 28th {ACM} Joint
  European Software Engineering Conference and Symposium on the Foundations of
  Software Engineering, Virtual Event, USA, November 8-13, 2020}},
  \bibfield{editor}{\bibinfo{person}{Prem Devanbu}, \bibinfo{person}{Myra~B.
  Cohen}, {and} \bibinfo{person}{Thomas Zimmermann}} (Eds.).
  \bibinfo{publisher}{{ACM}}, \bibinfo{pages}{788--799}.
\newblock
\urldef\tempurl%
\url{https://doi.org/10.1145/3368089.3409761}
\showDOI{\tempurl}


\bibitem[\protect\citeauthoryear{Zhang, Harman, Ma, and Liu}{Zhang
  et~al\mbox{.}}{2019}]%
        {DBLP:journals/corr/abs-1906-10742}
\bibfield{author}{\bibinfo{person}{Jie~M. Zhang}, \bibinfo{person}{Mark
  Harman}, \bibinfo{person}{Lei Ma}, {and} \bibinfo{person}{Yang Liu}.}
  \bibinfo{year}{2019}\natexlab{}.
\newblock \showarticletitle{Machine Learning Testing: Survey, Landscapes and
  Horizons}.
\newblock \bibinfo{journal}{\emph{CoRR}}  \bibinfo{volume}{abs/1906.10742}
  (\bibinfo{year}{2019}).
\newblock
\urldef\tempurl%
\url{http://arxiv.org/abs/1906.10742}
\showURL{%
\tempurl}


\bibitem[\protect\citeauthoryear{Zhang, Chen, Cheung, Xiong, and Zhang}{Zhang
  et~al\mbox{.}}{2018}]%
        {DBLP:conf/issta/ZhangCCXZ18}
\bibfield{author}{\bibinfo{person}{Yuhao Zhang}, \bibinfo{person}{Yifan Chen},
  \bibinfo{person}{Shing{-}Chi Cheung}, \bibinfo{person}{Yingfei Xiong}, {and}
  \bibinfo{person}{Lu Zhang}.} \bibinfo{year}{2018}\natexlab{}.
\newblock \showarticletitle{An empirical study on TensorFlow program bugs}. In
  \bibinfo{booktitle}{\emph{Proceedings of the 27th {ACM} {SIGSOFT}
  International Symposium on Software Testing and Analysis, {ISSTA} 2018,
  Amsterdam, The Netherlands, July 16-21, 2018}},
  \bibfield{editor}{\bibinfo{person}{Frank Tip} {and} \bibinfo{person}{Eric
  Bodden}} (Eds.). \bibinfo{publisher}{{ACM}}, \bibinfo{pages}{129--140}.
\newblock
\urldef\tempurl%
\url{https://doi.org/10.1145/3213846.3213866}
\showDOI{\tempurl}


\end{thebibliography}
\end{document}